%% file: main_arxiv.tex
\documentclass{article}

\usepackage[square,numbers]{natbib}
\usepackage[final]{neurips_2024}

\usepackage[utf8]{inputenc} % allow utf-8 input
\usepackage[T1]{fontenc}    % use 8-bit T1 fonts
\usepackage{hyperref}       % hyperlinks
\usepackage{url}            % simple URL typesetting
\usepackage{booktabs}       % professional-quality tables
\usepackage{amsfonts}       % blackboard math symbols
\usepackage{nicefrac}       % compact symbols for 1/2, etc.
\usepackage{microtype}      % microtypography
\usepackage{subfig}
\usepackage{graphicx} 
\usepackage{nicematrix}
\usepackage{tabularx}
\usepackage{booktabs}
\usepackage{multirow}
\usepackage{marvosym}

\usepackage{wrapfig}

\usepackage{algorithm}
\usepackage{algpseudocode}
\usepackage{pifont}
\newcommand{\circled}[1]{\textcircled{\raisebox{-0.6pt}{\footnotesize #1}}}
\algrenewcommand\algorithmicrequire{\textbf{Input:}}
\algrenewcommand\algorithmicensure{\textbf{Output:}}

\usepackage{xcolor, colortbl}
\definecolor{LightCyan}{rgb}{0.88,1,1}
\definecolor{Gray}{gray}{0.4}
\definecolor{LightCyan}{rgb}{0.82,0.82,1}
\definecolor{LightGray}{HTML}{E3E4E4}
\definecolor{Pearl}{HTML}{F1EAE3}%{FDEEF4}
\definecolor{yellow-small}{HTML}{FCF3CF}  
\definecolor{purple-med}{HTML}{E8DAEF}
\definecolor{green-large}{HTML}{A9DFBF}
\definecolor{tableblue}{HTML}{98AFC7}
\input{math_extension}

\title{DiPEx: Dispersing Prompt Expansion for Class-Agnostic Object Detection}

\author{%
Jia Syuen Lim\thanks{Equal Contribution} \quad Zhuoxiao Chen$^*$  \quad \textbf{Mahsa Baktashmotlagh} \\  \textbf{Zhi Chen}\quad \textbf{Xin Yu} \quad \textbf{Zi Huang} \quad \textbf{Yadan Luo}\thanks{Corresponding Author} \\
  The University of Queensland\\
  \texttt{\{jiasyuen.lim, zhuoxiao.chen, m.baktashmotlagh\}@uq.edu.au} \\ 
  \texttt{\{zhi.chen, xin.yu, helen.huang, y.luo\}@uq.edu.au} \\ 
}

\begin{document}

\maketitle

\begin{abstract}
  Class-agnostic object detection (OD) can be a cornerstone or a bottleneck for many downstream vision tasks. Despite considerable advancements in bottom-up and multi-object discovery methods that leverage basic visual cues to identify salient objects, consistently achieving a high recall rate remains difficult due to the diversity of object types and their contextual complexity.
  In this work, we investigate using vision-language models (VLMs) to enhance object detection via a self-supervised prompt learning strategy. Our initial findings indicate that manually crafted text queries often result in undetected objects, primarily because detection confidence diminishes when the query words exhibit semantic overlap. To address this, we propose a Dispersing Prompt Expansion (\textbf{DiPEx}) approach. DiPEx progressively learns to expand a set of distinct, non-overlapping hyperspherical prompts to enhance recall rates, thereby improving performance in downstream tasks such as out-of-distribution OD. Specifically, DiPEx initiates the process by self-training generic parent prompts and selecting the one with the highest semantic uncertainty for further expansion. The resulting child prompts are expected to inherit semantics from their parent prompts while capturing more fine-grained semantics. We apply dispersion losses to ensure high inter-class discrepancy among child prompts while preserving semantic consistency between parent-child prompt pairs. To prevent excessive growth of the prompt sets, we utilize the maximum angular coverage (MAC) of the semantic space as a criterion for early termination. We demonstrate the effectiveness of DiPEx through extensive class-agnostic OD and OOD-OD experiments on MS-COCO and LVIS, surpassing other prompting methods by up to 20.1\% in AR and achieving a 21.3\% AP improvement over SAM. The code is available at \href{https://github.com/jason-lim26/DiPEx}{https://github.com/jason-lim26/DiPEx}. 
\end{abstract}

\vspace{-1ex}
\section{Introduction}\label{sec:intro}
\vspace{-1ex}
In real-world applications, the class of interest may constantly change, prompting the need for new tasks like out-of-distribution (OOD) detection \cite{DBLP:conf/iccv/Wilson0DMS23,DBLP:conf/nips/DuGML22}, open-world detection \cite{DBLP:journals/tcsv/ZhaoMWSQL24,DBLP:conf/cvpr/00050CL0ZW23,DBLP:conf/cvpr/Joseph0KB21,DBLP:conf/cvpr/ZoharWY23,DBLP:conf/eccv/WuLCWKY22} and open-vocabulary \cite{DBLP:conf/cvpr/0001LDDLQCL23,DBLP:conf/cvpr/WuZ00L23,DBLP:conf/iclr/LinSJ0QHY023,DBLP:conf/iccv/LiMSTRYP23} object detection (OD) to ensure reliable operation of detectors. A significant bottleneck in these OD tasks is the ability to locate \textit{all} objects in a scene - typically referred to as class-agnostic OD \cite{DBLP:conf/eccv/MaazR0KA022}. Ensuring a high \textit{recall} rate is essential in this task as it lays the foundation for correctly classifying objects, thereby improving the average precision for classes of interest. Conversely, a low recall implies that some objects will be missed entirely, negatively impacting downstream recognition tasks.

Conventional solutions to the under-explored class-agnostic OD task often rely on bottom-up strategies \citep{DBLP:journals/ijcv/UijlingsSGS13, DBLP:conf/eccv/ZitnickD14, DBLP:conf/nips/PinheiroCD15, DBLP:journals/pami/Pont-TusetABMM17} such as selective search \cite{DBLP:journals/ijcv/UijlingsSGS13} or EdgeBox \cite{DBLP:conf/wacv/Jaiswal0NN21}, which generate a large ranked set of class-agnostic proposals based on low-level visual cues. To address the low precision and scalability issues of these approaches, another line of research has explored multi-object discovery by leveraging (self)-supervised features from vision transformers (ViT) (\textit{e.g.}, DINO \cite{DBLP:journals/corr/abs-2304-07193}, MoCo-v2 \cite{DBLP:journals/corr/abs-2003-04297}, SwAV \cite{DBLP:conf/nips/CaronMMGBJ20}), or external motion information to support region proposal regression. However, these methods still fall short, achieving only about 30\% average recall (AR) on benchmark datasets like MS-COCO due to the lack of intrinsic knowledge about a wide range of objects. The newly released vision-language models (VLMs) such as Grounding DINO \cite{liu2023grounding}, GLIP \cite{DBLP:conf/cvpr/LiZZYLZWYZHCG22}, T-Rex2 \cite{DBLP:journals/corr/abs-2403-14610}, which are pretrained on large-scale grounding datasets, have opened up new opportunities for acquiring common knowledge for generic object localization. VLMs have demonstrated impressive zero-shot recognition capacities given the provided \textit{textual prompt}. However, to effectively locate all objects, one would need to input all class names accurately, which is impractical in real-world applications.

\begin{figure}[t]
    \centering
    \vspace{-1ex}
    \subfloat[Class-agnostic OD and Downstream OOD-OD]{
        \includegraphics[width=0.5\textwidth]{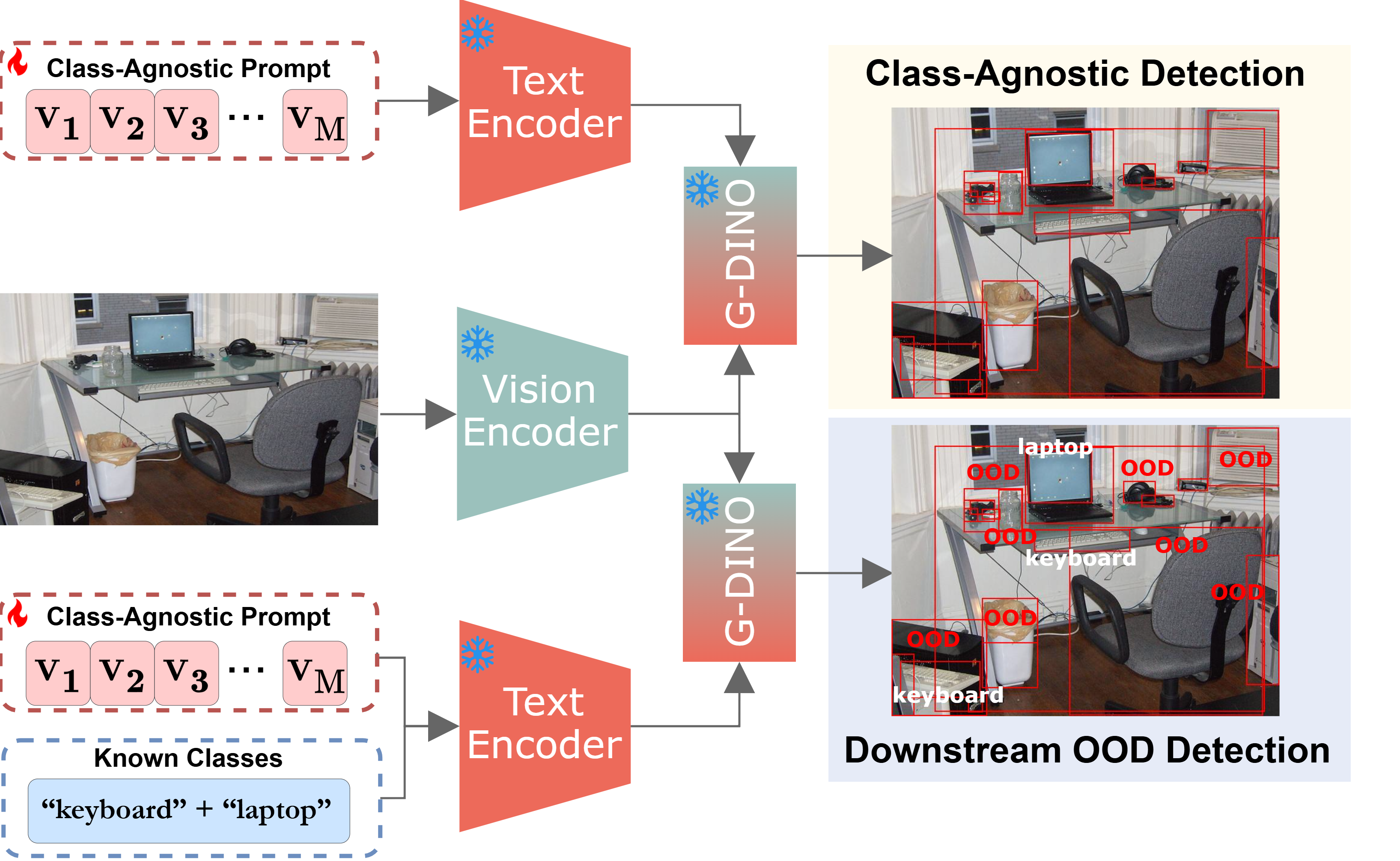}
        \label{fig:task}
    }%
    \subfloat[\textsc{Universal} and \textsc{Class-wide} Queries]{%
        \begin{minipage}[b]{0.48\textwidth}
            \centering
            \includegraphics[width=\textwidth]{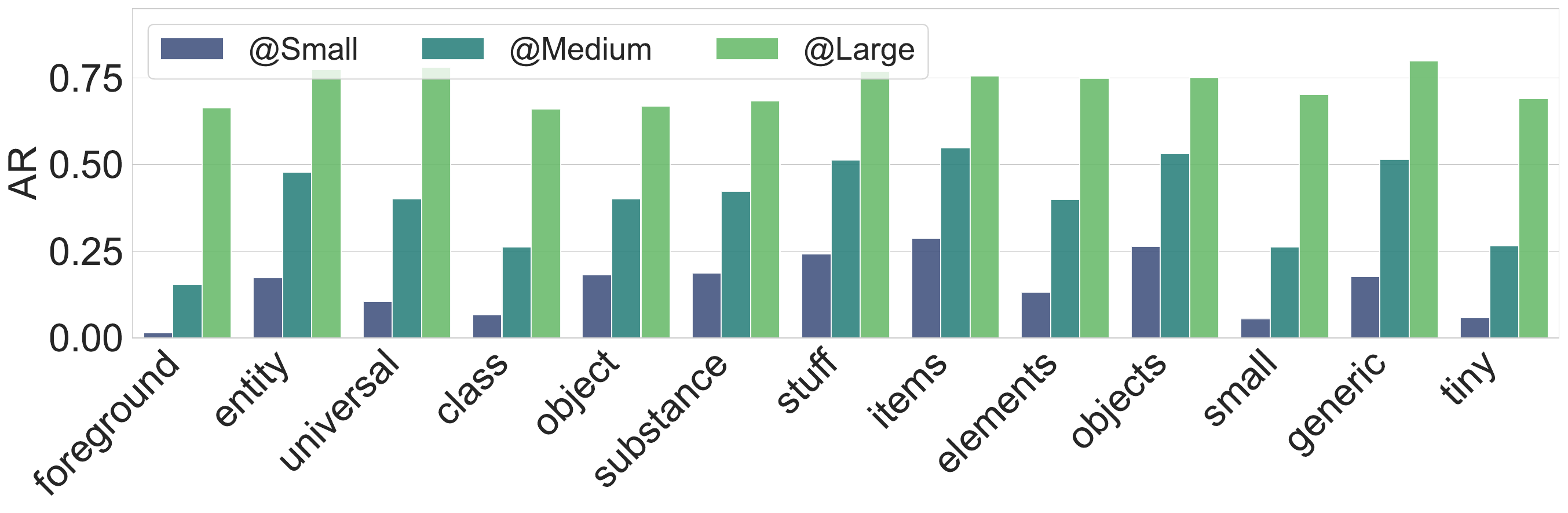}
            \vspace{-2ex}
            \includegraphics[width=\textwidth]{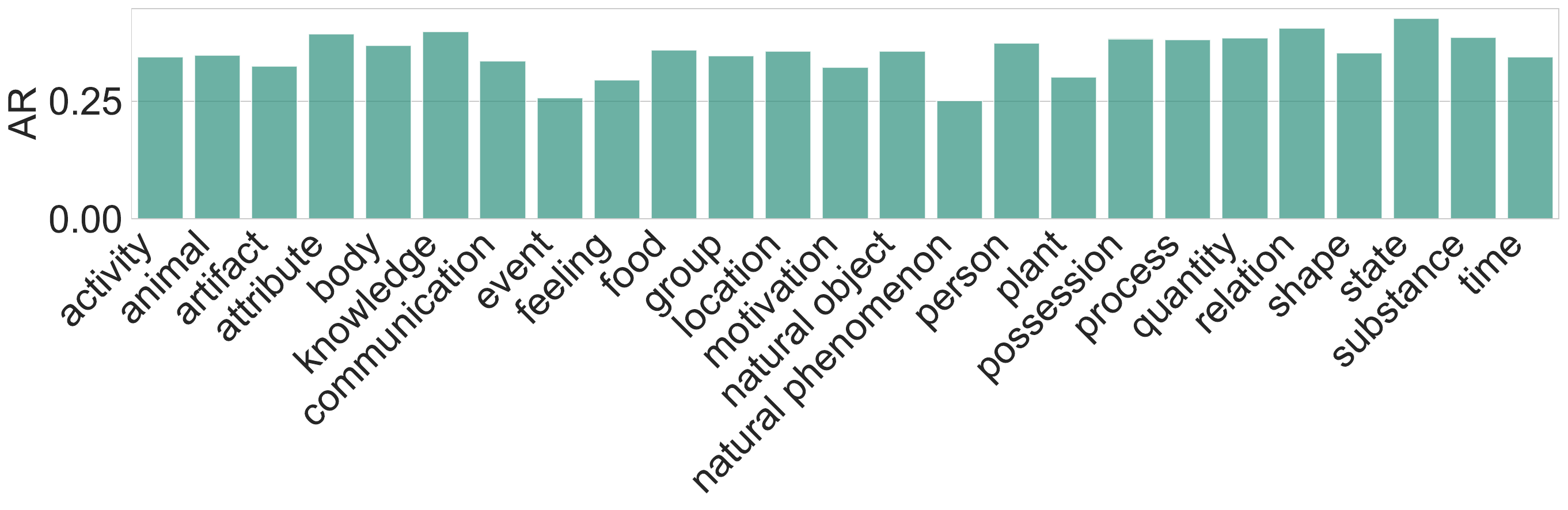}\vspace{-1.5ex}
            \label{fig:hand-crafted}
        \end{minipage}
    }%
    \caption{(a) An exemplar of the studied class-agnostic OD and downstream OOD-OD tasks. (B) Zero-shot class-agnostic OD performance of Grounding DINO \cite{liu2023grounding} on MS-COCO \cite{DBLP:conf/eccv/LinMBHPRDZ14}, with the hand-crafted \textsc{Universal} query from ChatGPT and \textsc{Class-wide} query from WordNet \cite{fellbaum1998wordnet}.\vspace{-3ex}}
    \label{fig:query}
\end{figure}

To better understand the limitation of modern VLMs in generic object localization, we investigated the design of hand-crafted text queries (Section \ref{sec:pilot})  to enhance detection recall through two approaches: (1) We employed a \textbf{\textsc{Universal} query,} using ChatGPT to generate 13 types of broad nouns and adjectives (\textit{e.g.}, “objects”, “generic”) as queries for the Grounding DINO model, aiming to detect a wide array of objects without focusing on specific categories; (2) We implemented a \textbf{\textsc{Class-wide} query}, selecting 25 high-level semantic words (\textit{e.g.}, “plant”, “animal”) from the top layer of the WordNet hierarchy (also used for the ImageNet vocabulary) to cover extensive object categories. Our findings, depicted in Figure \ref{fig:hand-crafted} and Table \ref{tab:concat}, reveal that while VLMs can generalize across universal object categories, combining all queries into \textit{one string} significantly reduces detection performance (by up to 52\% in AR) due to the ``\textbf{semantic overlap}'' among words. This suggests that optimal detection requires conducting multiple \textit{separate} inferences, presenting substantial computational demands for large datasets.

To overcome the aforementioned limitations, we propose a novel self-supervised Dispersing Prompt Expansion (DiPEx) strategy. This approach progressively expands a set of non-overlapping hyperspherical prompts for capturing all objects in a given dataset, thereby benefiting downstream tasks such as out-of-distribution object detection. Specifically, we start with a generic parent prompt that is self-supervised using the \textsc{Universal} and \textsc{Class-wide} text queries. To capture more fine-grained semantics, we split the parent prompts with high semantic uncertainty into a set of distinct child prompts. We initialize child prompts by diversifying the parent token embedding, randomly rotating it to different angles on the hypersphere to yield a range of unique prompts. Dispersion losses are employed to minimize semantic overlap among child prompts while maintaining semantic consistency across parent-child prompt pairs. To prevent excessive growth of the prompt sets, we estimate the maximum angular coverage (MAC) of the semantic space as a criterion to terminate the prompt expansion process, balancing semantic richness and computational overhead. Extensive experiments on the MS-COCO and LVIS datasets verify the effectiveness and versatility of the proposed DiPEx strategy. With a single pass of inference, DiPEx can achieve by up to 20.1\% improvements in average recall (particularly 35.2\% for small objects) and outperforms segment anything model (SAM) \cite{DBLP:conf/iccv/KirillovMRMRGXW23} by 21.3\% in average precision. 
% The source code is available at \href{https://github.com/jason-lim26/DiPEx}{https://github.com/jason-lim26/DiPEx}. 

\begin{table}[t]
    \centering
    \caption{Zero-shot class-agnostic object detection performance of Grounding DINO \cite{liu2023grounding} on MS-COCO \cite{DBLP:conf/eccv/LinMBHPRDZ14}, with \textit{hand-crafted} prompts from various sources. We report average recall (AR) and precision (AP) limited to a maximum of 100 detections per image. $\Delta$AR quantifies the percentage decrease in AR comparing “query-merging” to “prediction-merging” for forming \textit{multi-word} queries.}\label{tab:concat}
    \resizebox{1\linewidth}{!}{
    \begin{tabular}{@{}llccccccc@{}} % Adding an extra column for Delta AR (%)
        \toprule
        Word Source & Merging Strategy & AR & $\Delta$AR & AR@S & AR@M & AR@L & AP \\
        \midrule
        \multirow{2}{*}{ChatGPT \cite{DBLP:journals/corr/abs-2303-08774}}  & query-merging     & 0.345 & \cellcolor{Pearl!60} & 0.122 & 0.360 & 0.718 & 0.067 \\
                             & prediction-merging  & 0.526 & \multirow{-2}{*}{\cellcolor{Pearl!60}-52.46\%} & 0.317 & 0.606 & 0.781 & 0.274 \\
        \midrule
        \multirow{2}{*}{WordNet \cite{fellbaum1998wordnet}} & query-merging     & 0.461 & \cellcolor{Pearl!60} & 0.234 & 0.522 & 0.774 & 0.229 \\
                                 & prediction-merging & 0.570 & \cellcolor{Pearl!60}\multirow{-2}{*}{-23.64\%} & 0.382 & 0.646 & 0.796 & 0.344 \\
        \midrule
        \multirow{2}{*}{ChatGPT \cite{DBLP:journals/corr/abs-2303-08774}+WordNet \cite{fellbaum1998wordnet}} & query-merging  & 0.408 & \cellcolor{Pearl!60} & 0.162 & 0.471 & 0.751 & 0.121 \\
                                     & prediction-merging     & 0.589 & \cellcolor{Pearl!60}\multirow{-2}{*}{-44.36\%} & 0.410 & 0.665 & 0.798 & 0.353 \\
        \bottomrule
    \end{tabular}}\vspace{-2ex}
\end{table}

\noindent \textbf{Related Study.} The full discussions can be found in Section \ref{sec:related_work}. Traditional bottom-up approaches for region proposal generation, such as those by \cite{DBLP:journals/ijcv/UijlingsSGS13} and \cite{DBLP:conf/eccv/KrahenbuhlK14}, often face precision constraints despite high recall rates, limiting their scalability. Recent advancements in Vision Transformers (ViTs) by \cite{DBLP:conf/iccv/CaronTMJMBJ21} and \cite{DBLP:conf/iclr/DosovitskiyB0WZ21} have enabled self-supervised learning on massive datasets, extracting semantically meaningful features. Methods like LOST \cite{DBLP:conf/bmvc/SimeoniPVRGBPMP21} and TokenCut \cite{DBLP:conf/cvpr/Wang0H0CV22} use graph-based techniques but are limited to detecting a single object per image. MOST \cite{DBLP:conf/iccv/RambhatlaMCS23} addresses this with entropy-based box analysis but struggles with generalization. MAVL \cite{DBLP:conf/eccv/MaazR0KA022} uses a late fusion strategy with text queries, requiring full supervision and multiple inferences. Our approach eliminates the need for labels and achieves state-of-the-art performance with one-pass inference using non-overlapping prompts. Vision-Language Models (VLMs), like those by \cite{DBLP:conf/icml/RadfordKHRGASAM21} and \cite{DBLP:conf/icml/JiaYXCPPLSLD21}, have shown potential in learning generic concepts. HierKD \cite{DBLP:conf/cvpr/MaLGLCWZH22} and OV-DETR \cite{DBLP:conf/cvpr/ZareianRHC21} align image representations with captions and extend DETR to open-vocabulary settings. GLIP \cite{DBLP:conf/cvpr/LiZZYLZWYZHCG22}, Grounding DINO \cite{liu2023grounding}, and T-Rex2 \cite{DBLP:journals/corr/abs-2403-14610} integrate object detection and visual grounding. However, VLMs' effectiveness depends on textual cues, and prompt tuning, as introduced by CoOp \cite{zhou2022coop} and improved by CoCoOp and MaPLe \cite{DBLP:conf/cvpr/KhattakR0KK23}, offers a solution by optimizing soft prompts while keeping the model's parameters frozen. ProDA \cite{DBLP:conf/cvpr/LuLZL022} learns diverse prompts using a Gaussian model. DFKD-VLFM \cite{DBLP:conf/mm/XuanCYXLZ23} and PromptStyler \cite{DBLP:conf/iccv/ChoNKYK23} attempted to diversify a fixed number of prompts through contrastive approach. Despite these advancements, full supervision is typically required. UPL \cite{DBLP:journals/corr/abs-2204-03649} and POUF \cite{DBLP:conf/icml/TanwisuthZZHZ23} introduced unsupervised prompt learning, but adaptation for object detection remains limited. DiPEx is the first to apply prompt learning to class-agnostic object detection through a progressive self-training approach. 

\vspace{-2ex}
\section{Pilot Study} \vspace{-2ex}
In this section, we detail our preliminary exploration of the zero-shot detection capabilities using state-of-the-art VLM, Grounding DINO \cite{liu2023grounding}, to detect all objects irrespective of the associated classes on the MS-COCO dataset \cite{DBLP:conf/eccv/LinMBHPRDZ14} as illustrated in Figure \ref{fig:hand-crafted}. We conduct experiments using two types of text queries: \textbf{\textsc{Universal}} queries generated by ChatGPT for general object detection, and \textbf{\textsc{Class-Wide}} queries derived from WordNet, representing broad object categories. Our experiments reveal that semantic overlap between text queries impacts detection performance. To support this hypothesis, we conduct a case study showing that similar concatenated prompts reduce the model’s detection confidence.

\label{sec:pilot}\vspace{-1ex}
\subsection{Hand-crafted Queries for Class-agnostic Object Detection} \vspace{-1ex}
\noindent\textbf{\textsc{Universal} Query.} We employ ChatGPT to generate 13 synonyms of \textbf{universal} concepts, including nouns and adjectives, which are displayed as x-axis labels. The zero-shot object detection results, measured by average recall (AR) and precision (AP) across the top 100 confident boxes for each query text, are presented. The plot reveals that more general terms such as “generic” and “items” yield the highest  \textsc{AR}. Surprisingly, more specific descriptors like “foreground”, “small”, or “tiny” tend to reduce \textsc{AR} and do not effectively aid in identifying foreground or small objects.

\noindent\textbf{\textsc{Class-wide} Query.} We utilize 25 semantically independent beginner words (listed as x-axis labels in the bottom figure) from the highest level of the WordNet hierarchy \cite{fellbaum1998wordnet} as \textbf{class-wide} text queries. A variation in AR (0.26$\sim$0.43) is observed with different textual queries from WordNet, with a mean AR of 0.35. Compared to the mean AR of 0.37 across class-agnostic queries generated by ChatGPT, the zero-shot detection ability remains similar, regardless of the types of queries used.

\vspace{-1ex}
\noindent\textbf{Discussion on Multi-Word Queries.} 
The zero-shot results presented in Figure \ref{fig:hand-crafted} were obtained using \textit{single-word} prompts for the Grounding DINO. To explore whether combining \textit{multiple words} as prompts from a given source (\textit{e.g.}, WordNet) could improve zero-shot detection performance, we developed strategies for merging at both the input stage (query-merging) and the output stage (prediction-merging) as shown in Table \ref{tab:concat}. The query-merging strategy concatenates all input text queries (\textit{e.g.}, “foreground . elements . $\cdots$ tiny . objects .”) and performs a single-pass inference to obtain detections. The prediction-merging strategy, on the other hand, uses each text query individually for separate inference and then combines all box predictions.  Table \ref{tab:concat} shows that applying query-merging to \textsc{Universal} words results in a 52.46\% reduction in AR compared to prediction-merging, whereas \textsc{Class-wide} queries (\textit{e.g.}, from WordNet) achieve a smaller decrease in AR of only 23.64\%. These findings suggest that large semantic overlaps in concatenated queries (\textit{e.g.}, “stuff”, “objects” and “item” from ChatGPT) may greatly contribute to diminished object detection performance. To further investigate this phenomenon, we conducted a case study analyzing the impact of semantic overlap on detection performance, which is presented in the following section.

% Another case study to support this observation is also presented in Section \ref{sec:case_study}. Consequently, we hypothesize that if we could develop a method to learn a set of semantically \textit{non-overlapping} prompts for the target dataset, we could efficiently locate all objects with \textit{one-pass} inference using VLMs.
\vspace{-1ex}
\subsection{Confidence Diminishing when Text Query Semantically Overlap}
\label{sec:case_study} \vspace{-1ex}
\begin{figure}[t]\vspace{-1ex}
    {\includegraphics[width=0.99\textwidth]{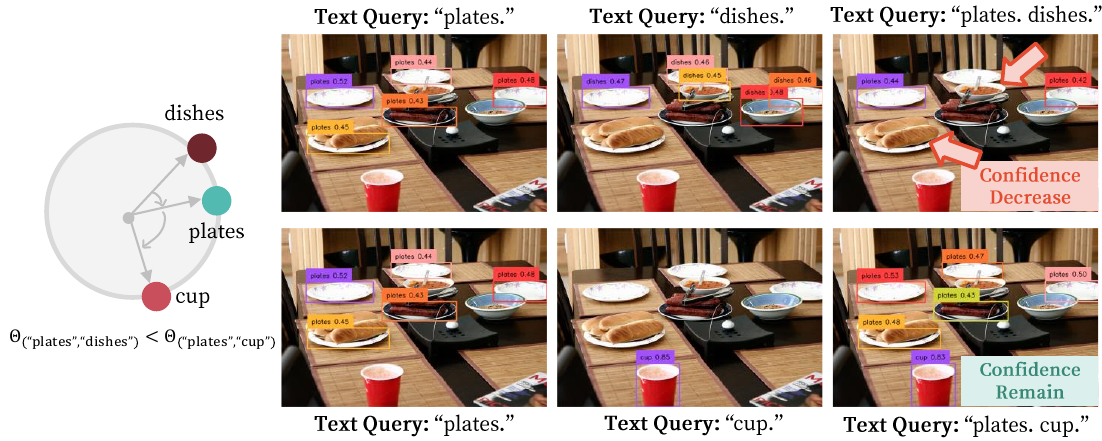}}\vspace{-1ex}
    \label{fig:kitti_results_boxes}%
    \caption{A case study investigating the impact of semantic overlap between text queries on the detection confidence of the pre-trained Grounding DINO \cite{liu2023grounding}. Semantic overlaps are quantified by the angular distance, denoted as $\Theta$, between tokenized embeddings of word pairs using BERT \cite{DBLP:conf/naacl/DevlinCLT19}. \vspace{-2ex}}\label{fig:pilot}
\end{figure}
To verify our hypothesis, we conduct a case study to demonstrate how semantic overlap in multi-word query leads to diminished detection confidence. We quantify semantic overlap by calculating the angular distance between pairs of textual token embeddings generated by BERT \cite{DBLP:conf/naacl/DevlinCLT19}. As shown in Figure \ref{fig:pilot}, a small angular distance $\theta$ of 53.73$^{\circ}$ between the text tokens \texttt{“plates”} and \texttt{“dishes”} diminishes the model's confidence. Consequently, some boxes that could be precisely localized with high confidence using the single token \texttt{“plates”} are omitted. In contrast, concatenating two text tokens with a larger angular distance (\textit{e.g.}, 60.99$^{\circ}$  between \texttt{“plates”} and \texttt{“cup”}) maintained high detection confidence. This combination resulted in bounding box predictions that encompassed all boxes predicted with each individual token (\texttt{“plates”} or \texttt{“cup”}). This case study supports our hypothesis that semantic overlap between concatenated text queries can interfere with the detection confidence of the model. Therefore, we propose that developing a method to learn a set of semantically \textit{non-overlapping} prompts for the target dataset could enable efficient object localization with \textit{one-pass} inference using VLMs.

\vspace{-1ex}
\section{Proposed Approach}\vspace{-1ex}
In this section, we first mathematically formulate the task of class-agnostic detection using a general VLM and, without loss of generality, illustrate the process using Grounding DINO \cite{liu2023grounding} as an exemplar model. We detail the steps of the proposed dispersing prompt expansion in Section \ref{sec:expansion}, followed by the early termination strategy of the prompt set growth.
\vspace{-1ex}
\subsection{Problem Formulation}\vspace{-1ex}
\noindent\textbf{Class-agnostic OD.} Let $\mathbf{I}$ denote the input image and $\mathbf{T}$ the associated text query. For the zero-shot object detection in a class-agnostic setting, we consider the text query  $\mathbf{T}$ to be of the form of “\texttt{a photo of a \{class\}}”, where the class token \texttt{\{class\}} is sampled from our predefined \textsc{Universal} (\textit{e.g.,} “objects”) or \textsc{Class-wide} (\textit{e.g.,} “plant”) sets as described in Section \ref{sec:pilot}. The text query is then tokenized and projected into word embeddings as $\mathbf{P} = \{\mathbf{v}_1, \mathbf{v}_2, \ldots, \mathbf{v}_M, \mathbf{c}\}$, where $\mathbf{v} = \{\mathbf{v}_i\}_{i=1}^{M}\in\mathbb{R}^{M\times d}$ indicates a set of $M$ contextual embeddings and $\mathbf{c}$ is the query text embedding. Here, $d$ indicates the dimensions of learnable tokens. The visual embeddings $\mathbf{E}_{v}$ extracted from the visual encoder and prompt embeddings $\mathbf{P}$ are fused jointly to prompt the VLM and generate the final bounding box predictions $\mathbf{O} = f(\mathbf{E}_{v}, \mathbf{P})\in \mathbb{R}^{N_B \times 4}$, with $f$ being the VLM, and $N_B$ being the number of predicted boxes. Formally, the objective of class-agnostic OD is to ensure that the generated bounding boxes can capture any objects as comprehensively as possible.

\textbf{Adapt Prompt Tuning for Class-agnostic OD.} Instead of relying on hand-crafted templates, prompt tuning approaches like CoOp \cite{zhou2022coop} and CoCoOp \cite{zhou2022cocoop}, originally developed for classification tasks, aim to \textit{learn} the context embeddings $\mathbf{v}$ with a frozen VLM using a supervised contrastive learning loss. To adapt these prompt learning approaches to the Grounding DINO \cite{liu2023grounding} detection framework, we first construct a pseudo label set $\mathcal{D}_{\operatorname{PSL}}$ from the zero-shot detection results with \textsc{Universal} and \textsc{Class-wide} text queries (see Section \ref{sec:more_ablation} for details). The prompt learning is then supervised by the standard box regression loss $\mathcal{L}_{\operatorname{box}}$, $\mathcal{L}_{\operatorname{giou}}$ and focal classification loss $\mathcal{L}_{\operatorname{cls}}$ as implemented in \cite{liu2023grounding}.

\vspace{-4ex}
\subsection{Dispersing Prompt Expansion (DiPEx) }\label{sec:expansion}\vspace{-3ex}
Unlike previous prompt tuning approaches, the proposed DiPEx strategy aims to iteratively grow a set of learnable prompts $\mathbf{P} = \{\mathbf{P}_{1}, \mathbf{P}_{2}, \ldots, \mathbf{P}_{L}\}$ in a tree hierarchy of depth $L$. To maximize the utility of prompts and ensure minimal semantic overlap among them, we assume $\mathbf{v}$ resides on the surface of a unit-hypersphere, \textit{i.e.,} $\|\mathbf{v}_i\|_2 = 1$. This assumption transforms the overlap minimization problem into maximizing the angular distances among the learned prompts. In the initial round, we set a single learnable parent prompt $\mathbf{P}_1 = \{\mathbf{v}\}$, which is self-trained using $\mathcal{D}_{\operatorname{PSL}}$ with the same procedure outlined above. In each subsequent round $l$ for $l\in [1, L]$, we identify the parent prompts of highest uncertainty and grow $K$ child prompts $\mathbf{P}_{l+1}\in\mathbb{R}^{K\times d}$ from it. The learned $\mathbf{v}^*_l$ is then frozen and stored in a parent queue $\mathbf{P}_{\operatorname{parent}}$. Prompt growth is terminated when the maximum angular coverage $\alpha_{\operatorname{max}}$ exceeds a certain threshold $\Tau_{\alpha}$.

\begin{figure}[t]
    \centering
    \includegraphics[width=1\linewidth]{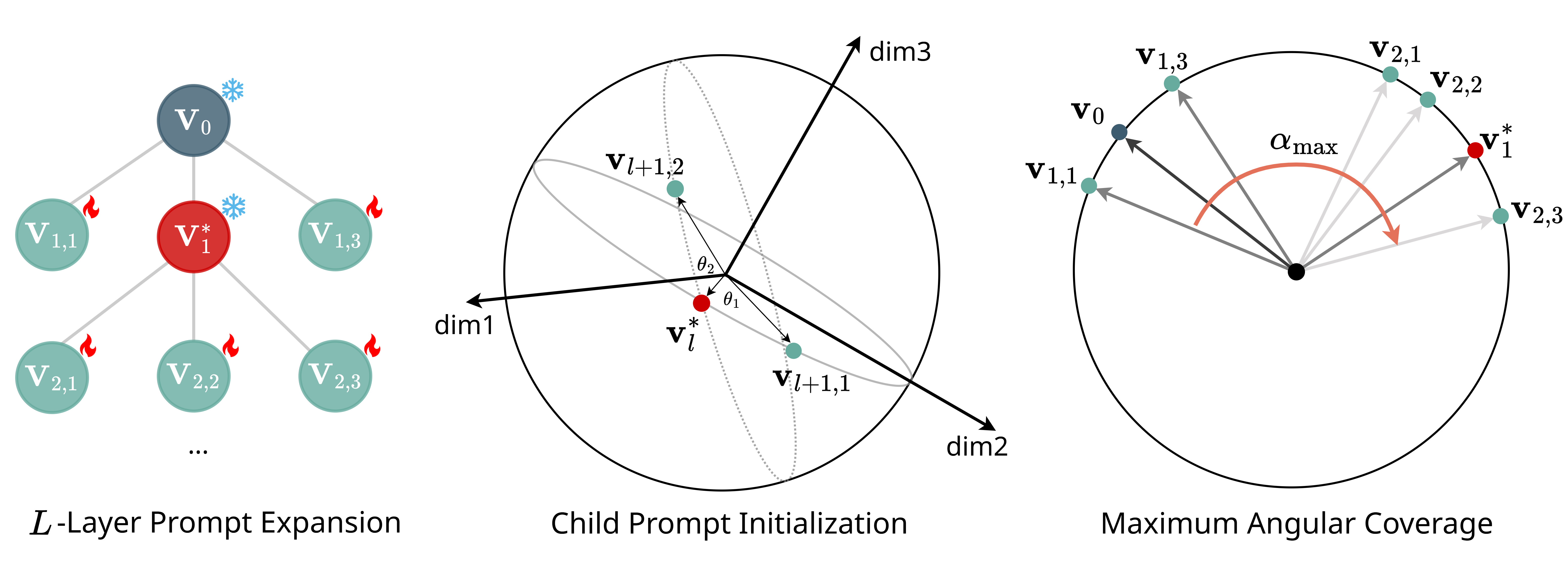}
    \caption{An illustration of the \circled{1} proposed prompt expansion strategy that selectively grows a set of child prompts for the highlighted parent prompt across $L$ iterations; \circled{2} diversifying initialized embeddings of the child prompt on a hypersphere and \circled{3} quantifying maximum angular coverage $\alpha_{\operatorname{max}}$ for early termination of the prompt growth.}\label{fig:dipex}\vspace{-5ex}
\end{figure}

\textbf{Child Prompt Initialization.} Continuing from the previous discussion, we now describe the process of child prompt initialization, which aims to inherit the semantics from parent prompts while capturing more fine-grained semantics. After the $l$-th round of training, we expand the parent prompt with the highest uncertainty, denoted as $\mathbf{v}^*_l\subset\mathbf{P}_l$, into a set of learnable child prompts (Figure \ref{fig:dipex}).  We empirically adopt the logit activation frequency of the prompts as a measure of uncertainty, visualized in Figure \ref{fig:activation}. The rationale is that if a prompt is activated for most samples, it covers overly broad semantics (\textit{e.g.,} animals) and may need to be decomposed into narrower categories (\textit{e.g.,} cats and dogs). To disentangle the complex semantic of $\mathbf{P}^*_l$, we set up $K$ child prompts $\mathbf{P}_{l+1} = \{\mathbf{v}_{l+1,k}\}_{k=1}^K$ for the selected parent prompt $\mathbf{v}^*_l$. To diversify the initialized embedding for each child prompt, we introduce $K$ random angular offsets $\mathbf{\Theta} = \{\mathbf{\theta}_{k}\}_{k=1}^K$ to rotate $\mathbf{v}^*_l$ on the hypersphere by different angles $\theta_k \sim[-\theta, \theta]$. Given that $\mathbf{v}^*_l$ is a $d$-dim vector, we randomly sample two axes $i$ and $j$ where $i,j\sim [1, d]$ for rotation. The $k$-th child prompt embedding $\mathbf{v}_{l+1,k}$ is then obtained by applying the corresponding  rotation matrix $\mathfrak{R}_k\in\mathbb{R}^{d\times d}$, which are defined as follows:
\begin{equation}\label{eq:child_prompt_init}
     \mathbf{v}_{l+1, k} =  \mathbf{v}_l^* \mathfrak{R}_{k}, \quad\mathfrak{R}_k = \begin{bsmallmatrix}
    1 &\scalebox{0.5}{$\cdots$} & 0 &\scalebox{0.5}{$\cdots$} & 0 & \scalebox{0.5}{$\cdots$} & 0 \\
    \scalebox{0.5}{$\vdots$} & \scalebox{0.5}{$\ddots$} & \scalebox{0.5}{$\vdots$} & & \scalebox{0.5}{$\vdots$} & & \scalebox{0.5}{$\vdots$} \\
    0 & \scalebox{0.5}{$\cdots$} & \cos \theta_k & \scalebox{0.5}{$\cdots$} & -\sin \theta_k & \scalebox{0.5}{$\cdots$} & 0 \\
    \scalebox{0.5}{$\vdots$} & & \scalebox{0.5}{$\vdots$} & 1 & \scalebox{0.5}{$\vdots$} & & \scalebox{0.5}{$\vdots$} \\
    0 & \scalebox{0.5}{$\cdots$} & \sin \theta_k & \scalebox{0.5}{$\cdots$} & \cos \theta_k & \scalebox{0.5}{$\cdots$} & 0 \\
    \scalebox{0.5}{$\vdots$} & & \scalebox{0.5}{$\vdots$} & & \scalebox{0.5}{$\vdots$} & \scalebox{0.5}{$\ddots$} & \scalebox{0.5}{$\vdots$} \\
    0 & \scalebox{0.5}{$\cdots$} & 0 & \scalebox{0.5}{$\cdots$} & 0 & \scalebox{0.5}{$\cdots$} & 1
    \end{bsmallmatrix}.
\end{equation}

Here, the non-identity elements are placed at the intersections of the \( i \)-th and \( j \)-th rows and columns, corresponding to the plane of rotation, as illustrated by the grey ellipses in Figure \ref{fig:dipex}. As the initialized embeddings of the child prompts are diversified while maintaining consistency with the central parent embedding (red dot), this leads to varying detection results. This enriched prediction diversity allows us to facilitate online self-training, where we adopt the predictions with the highest confidence as pseudo labels for each child prompt, which in turn supervise the next round of prompt learning with respect to $\mathcal{L}_{\operatorname{bbox}}$, $\mathcal{L}_{\operatorname{giou}}$ and $\mathcal{L}_{\operatorname{cls}}$ for the next iteration.

\textbf{Optimization.} We expect the learned child prompts to follow an accurate semantic hierarchy, having minimal overlap with other child tokens while maintaining semantic consistency with their original parent prompts. We leverage the following \textit{dispersion losses} to enlarge the angular distances among the child-child and decrease the distances between child-parent prompt pairs:\vspace{-1ex}
\begin{equation}
\begin{split}
    \mathcal{L}_{\operatorname{parent-child}} &= -\frac{1}{K} \sum_{i=1}^{K}(\frac{\mathbf{v}_i^{\top}\mathbf{v}^*_l}{\|\mathbf{v}_i\|\|\mathbf{v}^*_l\|}/ \tau_p),\\ \mathcal{L}_{\operatorname{child-child}} &= \frac{1}{K} \sum_{i=1}^{K}\log\frac{1}{K - 1}\sum_{j\neq i}\exp(\frac{\mathbf{v}_i^{\top}\mathbf{v}_j}{\|\mathbf{v}_i\|\|\mathbf{v}_j\|} / \tau_c),\vspace{-1ex}
\end{split}
\end{equation}
where $\mathbf{v}_l^*$ is retrieved from the parent prompt queue $\mathbf{P}_{\operatorname{parent}}$ as a fixed prototype. The temperature coefficients $\tau_p$ and $\tau_c$ adjust the angular separation. The overall optimization can be formulated as:
\begin{equation}\label{eq:train_prompts}
    \mathcal{L} = \mathcal{L}_{\operatorname{parent-child}} + \gamma \mathcal{L}_{\operatorname{child-child}} + \gamma_{\operatorname{bbox}}\mathcal{L}_{\operatorname{bbox}} + \gamma_{\operatorname{giou}}\mathcal{L}_{\operatorname{giou}} + \gamma_{\operatorname{cls}}\mathcal{L}_{\operatorname{cls}},
\end{equation}
where $\gamma$ is the loss coefficient that controls the $\mathcal{L}_{\operatorname{child-child}}$. The rest coefficients i.e., $\gamma_{\operatorname{bbox}}, \gamma_{\operatorname{giou}}, \gamma_{\operatorname{cls}}$ follows \cite{liu2023grounding}. Until the optimization convergence, the prompt expansion will repeat if needed.

\noindent\textbf{Expansion Termination with Maximum Angular Coverage 
 (MAC)}. While prompt expansion is effective in capturing fine-grained semantics, it inevitably introduces computational overhead, impacting inference efficiency for downstream tasks. To balance the semantic richness and inference costs, we gather all learned prompts $\mathbf{P}$ and evaluate the maximum angular coverage (MAC) among all pairs. MAC is defined as:
\begin{equation}\label{eq:mac}
    \alpha_{\operatorname{max}} = \max_{\mathbf{v}_i, \mathbf{v}_j\in \mathbf{P}} \operatorname{arccos}(\frac{\mathbf{v}_i^{\top}\mathbf{v}_j}{\|\mathbf{v}_i\|\|\mathbf{v}_j\|}).
\end{equation}
The $\alpha_{\operatorname{max}}$ reveals the breadth of vocabularies covered by the current prompts. Notably, our empirical study shows that as the number of expansion rounds increases, the MAC increases monotonically and eventually converges. This convergence serves as an effective \textit{signal} to terminate prompt expansion. The overall algorithm is summarized in Algorithm \ref{alg:dipex}.

\begin{algorithm}[t]
\caption{The Proposed DiPEx for Class-Agnostic Object Detection}
\begin{algorithmic}
\Require $f$: vision-language model (VLM)
\Ensure $\mathbf{P}$: set of fine-tuned prompts for $f$ to detect class-agnostic objects
\State Initialize a single learnable parent prompt $\mathbf{P}_1 = \{\mathbf{v}_1\}$
\State Optimize $\mathbf{P}_1$ using zero-shot detection results from $f$
\State Initialize a growing set of learnable prompts $\mathbf{P} = \{\mathbf{P}_1\}$ and an empty parent queue $\mathbf{P}_{\operatorname{parent}}=\{\}$
\For{each round $l \in \{1, 2, \cdots, L\}$}
    \State Identify the parent prompt with the highest uncertainty $\mathbf{v}^*_l \in \mathbf{P}_l$
    \State Freeze $\mathbf{v}^*_l$ and add it to the parent queue $\mathbf{P}_{\operatorname{parent}}$
    \State Expand $\mathbf{v}^*_l$ into $K$ learnable child prompts $\mathbf{P}_{l+1} = \{\mathbf{v}_{l+1,k}\}_{k=1}^K$ via Equation \eqref{eq:child_prompt_init}
    \State Grow the set of learnable prompts: $\mathbf{P} = \mathbf{P}_{l} \cup \mathbf{P}_{l+1}$
    \State Optimize the prompts in $\mathbf{P}$ using Equation \eqref{eq:train_prompts} with $\mathbf{P}_{\operatorname{parent}}$
    \State Compute maximum angular coverage (MAC) via Equation \eqref{eq:mac}
    \If{MAC converges}
        \State Break; terminate the prompt growth
    \EndIf
\EndFor
\end{algorithmic}
\label{alg:dipex}
\end{algorithm}

\vspace{-1ex}
\section{Experiments}\vspace{-1ex}
\subsection{Experimental Setup}\vspace{-1ex}
\noindent \textbf{Datasets.} We conduct our experiments using two detection datasets: 1). \textbf{MS-COCO} \cite{DBLP:conf/eccv/LinMBHPRDZ14}, a large-scale object detection and instance segmentation dataset, comprising approximately 115K training images and 5K validation images across 80 classes. 2). \textbf{LVIS} \cite{DBLP:conf/cvpr/GuptaDG19} includes 2.2 million high-quality instance segmentation masks covering 1,000 class labels, resulting in a long-tailed data distribution. It consists of around 100K training images and 19.8K validation images. For class-agnostic object detection (CA-OD) setting, we merge all categories from both datasets into a single class to perform class-agnostic detection. To further validate the efficacy of DiPEx in downstream out-of-distribution object detection (OOD-OD) tasks, we evaluate our method using a rectified version of the OOD-OD benchmark. Unlike previous benchmarks \cite{DBLP:conf/iclr/DuWCL22}, where samples that do not contain ID instances are manually selected and ID and OOD performance are evaluated separately, we tested our approach on the MS-COCO, which includes a mixture of both ID and OOD objects. While we followed the settings outlined in OOD-OD \cite{DBLP:conf/iclr/DuWCL22}, with 20 base classes in PASCAL-VOC \cite{DBLP:journals/ijcv/EveringhamGWWZ10} designated as ID classes and the remaining classes treated as OOD. Our choice of dataset enhances the rigor of our evaluation by combining both ID and OOD instances, providing a more realistic assessment of our method's real-world conditions.

\textbf{Evaluation Metrics.} We report results for class-agnostic object detection on both the MS-COCO and LVIS validation splits. For evaluation, we adopt official metrics from the COCO 2017 challenge. Specifically, we report average precision (AP) at IoU thresholds from 0.5 to 0.95, along with average recall (AR) across the same threshold range. We also report AR by object scale: AR@S for small, AR@M for medium, and AR@L for large objects. Details on our implementation, including those of prior works used as \textit{baselines}, are provided in Appendix \ref{sec:impl_details}.

\begin{table*}[t]
\centering 
\caption{\textbf{Class-agnostic object detection} on the MS-COCO dataset. [ ] indicate the prompt word for Grounding DINO. The prompting methods indicated with `*' are adapted to the OD task.}
\vspace{-1ex}
\resizebox{1\linewidth}{!}{%
\begin{tabular}{l l c c c c c c c}
\toprule
Method & Description & AR$_{1}$ & AR$_{10}$ & AR$_{100}$ & AR@S & AR@M & AR@L &  AP \\
\midrule 
\midrule
Selective Search \cite{DBLP:journals/ijcv/UijlingsSGS13} & non-parametric & 0.1 & 1.1 & 7.8 &0.9 &7.2 &20.7 &0.1 \\
UP-DETR \cite{DBLP:conf/cvpr/DaiCLC21} & self-training & 0.2 & 1.4 & 1.4 & 0.0 & 0.2 & 5.8 & 0.1 \\
DETReg \cite{DBLP:conf/cvpr/BarWKRHCRDG22} & self-training & 0.6 & 3.7 & 12.9 & 0.2 & 12.8 & 35.3 & 1.4 \\
FreeSOLO \cite{DBLP:conf/cvpr/WangYMKASA22} & self-training & 3.7 & 9.7 & 12.6 & 0.5 & 12.3 & 34.1 & 4.2 \\
Exemplar-FreeSOLO \cite{DBLP:conf/cvpr/IshtiakEG23} & self-training & 8.2 & 13.0 & 17.9 & \textcolor{Gray}{--} & \textcolor{Gray}{--} & \textcolor{Gray}{--} & 12.6 \\
MOST \cite{DBLP:conf/iccv/RambhatlaMCS23} & self-training & 3.1 & 6.4 & 6.4 & 0.1 & 1.6 & 24.5 & 3.3 \\
CutLER \cite{DBLP:conf/cvpr/0007GYM23} & self-training & 6.8 & 19.6 & 32.8 & 13.7 & 37.5 & 60.0 & 29.6 \\
\midrule
Grounding DINO [“generic”] \cite{DBLP:conf/naacl/DevlinCLT19} & zero-shot & 10.3 & 37.8 & 44.1 & 17.7 & 51.6 & 80.0 & 28.3 \\
Grounding DINO+CoOp$^*$ \cite{zhou2022coop} & self-training & 10.4 & 39.1 & 61.3 & 36.4 & 72.7 & 88.8 & 34.6 \\
Grounding DINO+CoCoOp$^*$ \cite{zhou2022cocoop} & self-training & 7.6 & 34.1 & 58.1 & 33.9 & 68.3 & 86.1 & 24.6 \\

  \rowcolor{LightCyan!40}\textbf{DiPEx} & self-training &\textbf{10.5} &\textbf{40.8} &\textbf{63.2} &\textbf{39.2 } &\textbf{74.3} &\textbf{89.8} &\textbf{35.9} \\
\bottomrule
\end{tabular}}\vspace{-2ex}
\label{table:uod_coco}
\end{table*}

\subsection{Main Results on Class-agnostic OD and OOD-OD}\vspace{-1ex}
\textbf{Class-agnostic OD on MS-COCO.} To validate our proposed method for class-agnostic object detection, we compared it against ten different baseline methods on the MS-COCO dataset, using various metrics as reported in Table \ref{table:uod_coco}. We observed that non-parametric methods generally underperform compared to self-training methods due to their inability to learn and extract semantic and geometric information about objects from the dataset. In contrast, Grounding DINO, leveraging pre-trained knowledge, demonstrates strong zero-shot capabilities and achieves AR$_{100}$ of 44.1\% with a single text prompt, ``generic''. Furthermore, CoOp, which fine-tunes prompts for Grounding DINO, enhances class-agnostic detection performance by 39.0\% in AR$_{100}$ compared to direct zero-shot inference. Our method, which expands the learnable prompts to a wider angular distance, surpasses all baselines by achieving the highest performance across all metrics and outperforming the leading baseline, CoOp, by 3.1\% in AR$_{100}$. Notably, for \textit{small objects} which are challenging to localize, our method improves AR@S by 7.7\% compared to CoOp, indicating that expanded prompts better capture a range of object sizes. Additionally, the proposed DiPEx achieved the highest AP of 35.9\%, demonstrating the superior quality of class-agnostic detection.

\textbf{Class-agnostic OD on LVIS.}  
To further validate the efficacy of DiPEx, we conducted extensive experiments on the challenging LVIS dataset, which includes thousands of classes with a long-tail distribution. As shown in Table \ref{table:uod_lvis}, prompt tuning methods such as CoOp \cite{zhou2022coop} and CoCoOp \cite{zhou2022cocoop} outperform zero-shot Grounding DINO when using hand-crafted prompts (\textit{e.g.}, “items”, “generic”, “objects”). Additionally, CoCoOp surpasses multi-object discovery baselines like CutLER \cite{DBLP:conf/cvpr/0007GYM23} and HASSOD \cite{DBLP:conf/nips/CaoJGW23}, by 86.7\% and 51.3\% in AR$_{200}$, respectively. Notably, SAM \cite{DBLP:conf/iccv/KirillovMRMRGXW23}, which was pre-trained on a vast of dataset containing millions of images and billions of masks, demonstrates strong zero-shot capabilities, surpassing all other baselines. In contrast, our proposed DiPEx outperforms SAM by 13.3\% in AR$_{200}$ and 21.3\% in AP after only four epochs of self-training, Furthermore, DiPEx exceeds CoOp by 20.1\% in AR$_{200}$.

\begin{table*}[t]
\centering 
\caption{\textbf{Class-agnostic object detection} on the LVIS dataset. \textsuperscript{\textdagger} indicate the model is fine-tuned on the LVIS training set by self-training without box annotations. \vspace{-1ex}}
\resizebox{1\linewidth}{!}{%
\begin{tabular}{l c c c c c c c c c c}
\toprule
\multirow{1}{*}{Method} & AR$_{1}$ & AR$_{10}$ & AR$_{200}$ & AR@S & AR@M & AR@L & AP & AP@S & AP@M & AP@L \\
\midrule 
Selective Search \cite{DBLP:journals/ijcv/UijlingsSGS13} & 0.1 & 1.1 & 13.0 & 6.1 & 19.9 & 37.6 & 0.2 & 0.4 & 0.2 & 0.2 \\
G-DINO [“object”] \cite{DBLP:conf/naacl/DevlinCLT19} & 4.1 & 17.9 & 27.2 & 13.0 & 44.1 & 71.1 & 5.4 & 5.6 & 10.0 & 9.4 \\
G-DINO [“generic”] \cite{DBLP:conf/naacl/DevlinCLT19} & 3.8 & 16.5 & 20.2 & 6.5 & 34.5 & 67.7 & 9.0 & 4.1 & 17.4 & 30.7 \\
G-DINO [“items”] \cite{DBLP:conf/naacl/DevlinCLT19} & 4.0 & 17.8 & 28.0 & 13.9 & 45.3 & 70.7 & 11.6 & 6.3 & 19.6 & 32.0 \\
SAM \cite{DBLP:conf/iccv/KirillovMRMRGXW23} & \textcolor{Gray}{--} & \textcolor{Gray}{--} & 42.7 & 27.7 & 66.3 & 75.5 & 6.1 & \textcolor{Gray}{--} & \textcolor{Gray}{--} & \textcolor{Gray}{--} \\
\midrule
\textsuperscript{\textdagger} CutLER \cite{DBLP:conf/cvpr/0007GYM23} & 2.4 & 9.3 & 21.8 & 10.8 & 35.1 & 55.5 & 4.5 & 2.7 & 9.1 & 15.1 \\
\textsuperscript{\textdagger} HASSOD \cite{DBLP:conf/nips/CaoJGW23} & 0.2 & 10.6 & 26.9 & 15.6 & 42.2 & 56.9 & 4.9 & 2.8 & 7.9 & 12.2 \\
\textsuperscript{\textdagger} G-DINO + CoOp$^*$ \cite{zhou2022coop} & 4.2 & 19.1 & 40.3 & 23.6 & 63.5 & 83.5 & 14.0 & 8.3 & 23.7 & 32.3 \\
\textsuperscript{\textdagger} G-DINO + CoCoOp$^*$ \cite{zhou2022cocoop} & 4.2 & 19.2 & 40.7 & 24.1 & 63.8 & 84.1 & 13.6 & 8.1 & 22.4 & 30.1 \\
\midrule
\rowcolor{LightCyan!40}\textsuperscript{\textdagger} \textbf{DiPEx} & \textbf{4.3} & \textbf{20.1} & \textbf{48.4} & \textbf{31.9} & \textbf{72.6} & \textbf{88.2} & \textbf{15.2} & \textbf{9.3} & \textbf{25.3} & \textbf{32.8} \\
\bottomrule
\end{tabular}}
\vspace{-1ex}
\label{table:uod_lvis}
\end{table*}

\begin{table*}[t]
\centering 
\caption{The downstream \textbf{out-of-distribution object detection (OOD-OD)} on the MS-COCO dataset, where the ground truth boxes contain both known and unknown classes.}\vspace{-1ex}
\resizebox{0.99\linewidth}{!}{%
\begin{tabular}{l c c c c c c c c c c}
\toprule
\multirow{2}{*}{Method} & \multicolumn{2}{c}{\cellcolor{Pearl!50}\textsc{Known}} & \multicolumn{8}{c}{\cellcolor{orange!10}\textsc{Unknown}} \\
\cmidrule(lr){2-3} \cmidrule(lr){4-11}
 & AP & AP50 & AR$_{100}$ & AR@S & AR@M & AR@L &AP &AP@S &AP@M &AP@L\\
\midrule 
Selective Search \cite{DBLP:journals/ijcv/UijlingsSGS13} & \textcolor{Gray}{--} & \textcolor{Gray}{--} & 8.3 & 1.0 & 8.5 & 23.2 & 0.1 & 0.0 & 0.0 & 0.5 \\
MOST \cite{DBLP:conf/iccv/RambhatlaMCS23}& \textcolor{Gray}{--} & \textcolor{Gray}{--} & 5.3 & 0.1 & 1.3 & 22.5 & 0.4 & 0.1 & 0.4 & 1.2 \\
CutLER \cite{DBLP:conf/cvpr/0007GYM23} & \textcolor{Gray}{--} & \textcolor{Gray}{--} & 34.5 & 15.8 & 41.5 & 62.7 & 5.7 & 2.3 & 6.9 & 13.7 \\
VOS \cite{DBLP:conf/iclr/DuWCL22}& 36.6 & 56.7 & 10.0 & 2.2 & 6.1 & 27.1 & 2.8 & 0.8 & 2.2 & 7.2 \\
PROB  \cite{DBLP:conf/cvpr/ZoharWY23} & 28.2 & 43.8 & 13.2 & 1.9 & 11.2 & 40.3 & 0.9 & 0.6 & 0.9 & 2.1 \\
UnSniffer \cite{DBLP:conf/cvpr/LiangXLZM23} & 35.8 & 55.8 & 20.6 & 11.8 & 19.9 & 34.8 & 2.9 & 1.5 & 3.1 & 5.3 \\
G-DINO [“generic”] & \textbf{46.3} & \textbf{59.7} & 43.3 & 18.0 & 52.1 & 82.6 & 12.5 & 6.9 & 17.8 & \textbf{25.7} \\
\midrule
  \rowcolor{LightCyan!40}\textbf{DiPEx} & \textbf{46.3} & \textbf{59.7} & \textbf{59.9} & \textbf{35.8} & \textbf{72.9} & \textbf{89.7} & \textbf{15.7} & \textbf{9.7} & \textbf{21.8} & 25.2 \\
\bottomrule
\end{tabular}}
\vspace{-2ex}
\label{table:ood_coco}
\end{table*}

\begin{figure}[t]
    \centering
    \vspace{-1ex}
    \includegraphics[width=0.49\textwidth]{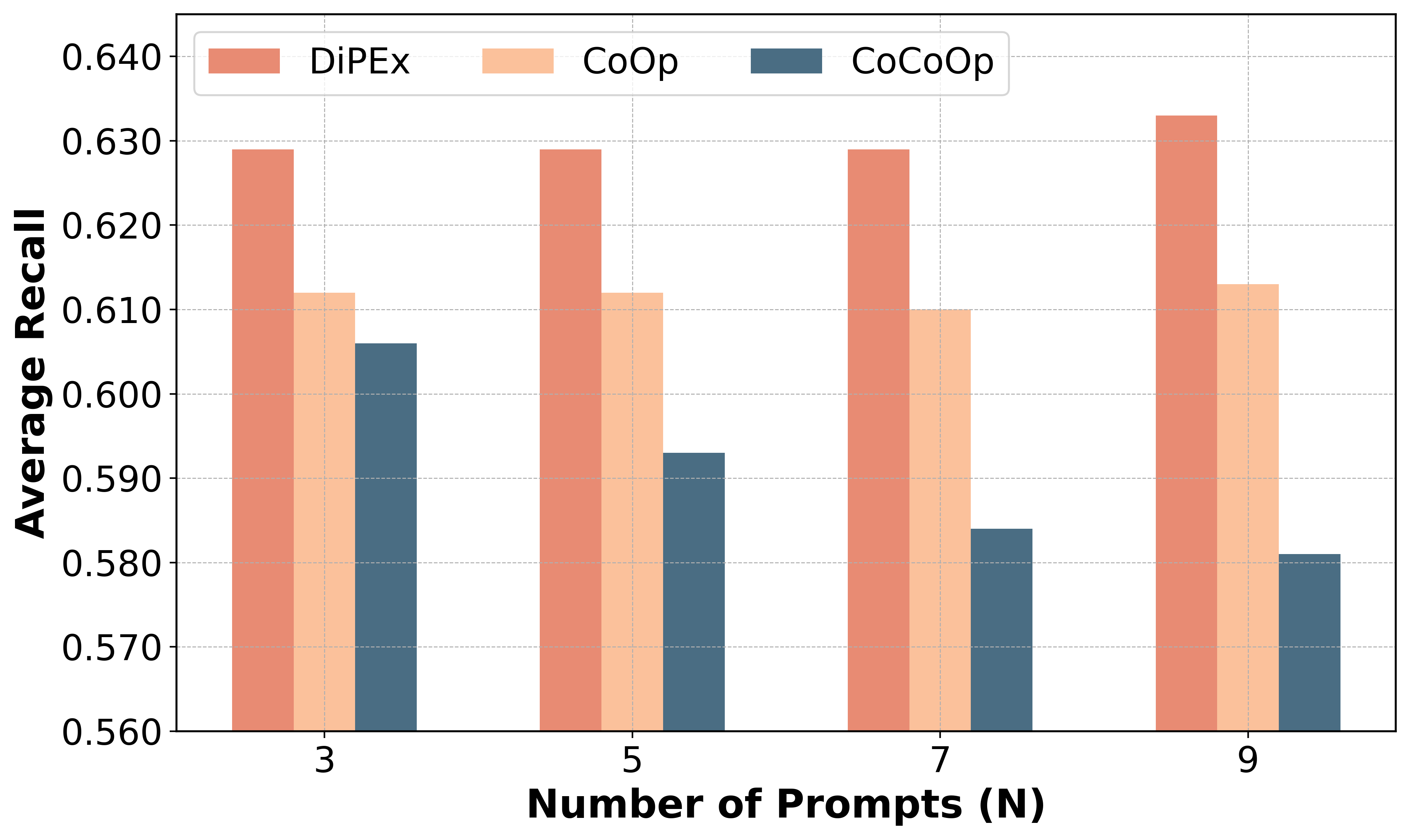}\hfill
    \includegraphics[width=0.49\textwidth]{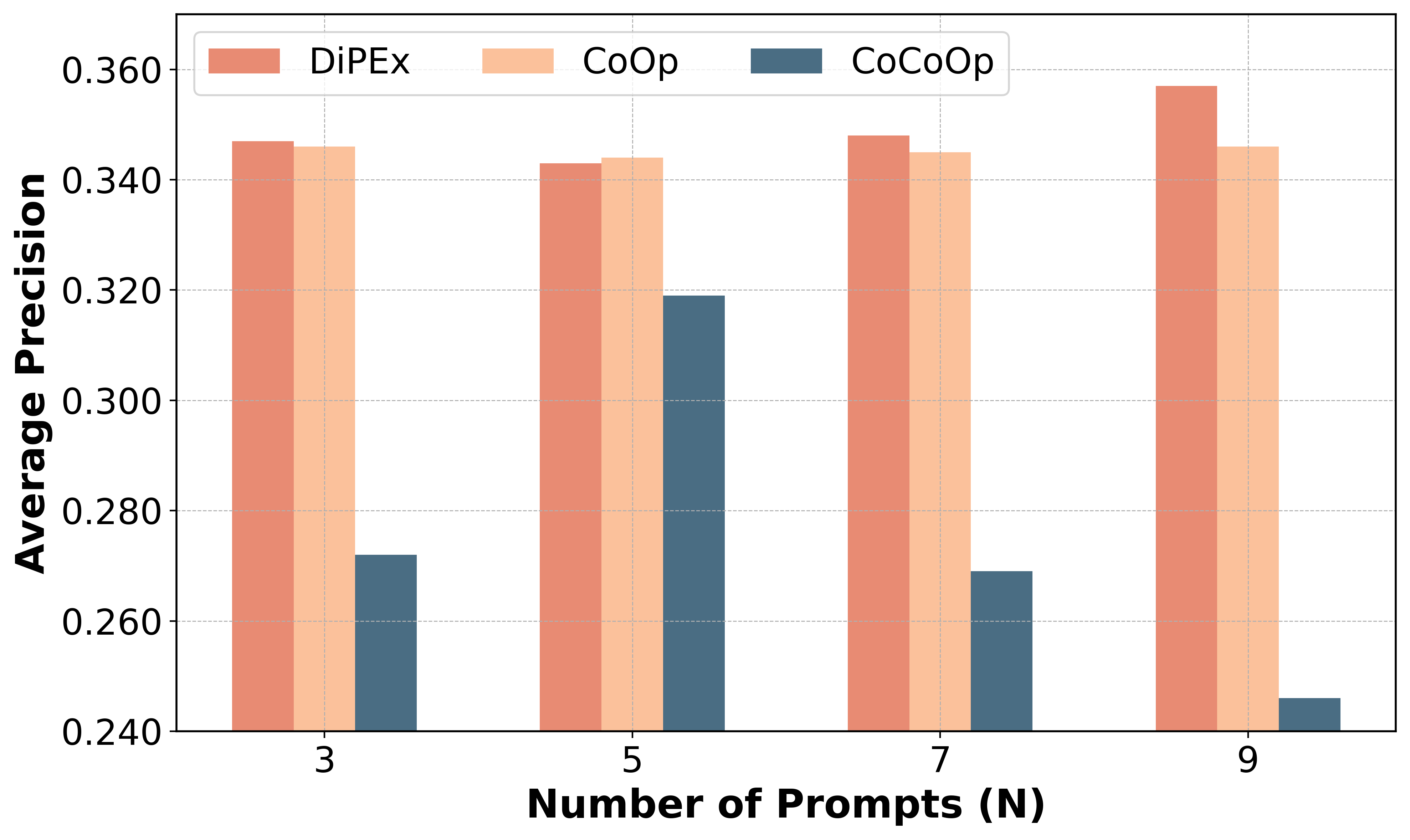}
    \caption{Impact of the prompt length on the MS-COCO dataset. The average recall (AR) and precision (AP) are reported to compare the derived DiPEx against CoOp \citep{zhou2022coop} and CoCoOp \cite{zhou2022cocoop}.}
    \label{fig:prompt_length}
\end{figure}

\textbf{Downstream OOD-OD on MS-COCO.} To evaluate the generalization of our proposed DiPEx in out-of-distribution object detection (OOD-OD), we compared its performance on both known and unknown classes against various baselines. As shown in Table \ref{table:ood_coco}, the zero-shot Grounding DINO uses known class names as prompts, supplemented with a simple “generic” prompt for unknowns, outperforms all other non-VLM methods (\textit{e.g.}, 25.5\% higher AR$_{100}$ compared to CutLER \cite{DBLP:conf/cvpr/0007GYM23}). This improvement stems from VLMs leveraging rich semantic knowledge from language models to better comprehend object information in images. DiPEx enhances this further by expanding text prompts in embedding space, enabling it to capture and differentiate objects of varying sizes and diverse semantics from learned classes. This approach delivers a significant performance gain, achieving a 38.3\% increase in AR$_{100}$ and a 25.6\% in AP increase over zero-shot predictions. Furthermore, the expanded prompts can be directly applied alongside various known class vocabularies to detect unknown objects, eliminating the need for retraining.

\vspace{-2ex}
\subsection{Ablation Study and Model Analysis}
\vspace{-1ex}
We investigate the impact of various factors on prompting performance including the learnable prompt lengths, the number of expansion rounds $L$, and angular coverage achieved across rounds. To facilitate model analysis, we present the distribution of prompt logit activation and visualization of detection results. Further ablation studies refers to Section \ref{sec:more_ablation}.

\textbf{Impact on Number of Prompts.} In Figure \ref{fig:prompt_length}, we compare the impact of prompt length $N$ for DiPEx against CoOp \cite{zhou2022coop} \& CoCoOp \cite{zhou2022cocoop}. Overall, DiPEx shows consistent improvement in performance with a greater number of prompts -- not merely due to quantity, but rather because a larger set fosters greater diversification, enabling the model to capture more comprehensive semantics. In contrast, CoOp's \cite{zhou2022coop} performance remains constant, while CoCoOp's \cite{zhou2022cocoop} performance declines, suggesting that more prompts do not necessarily guarantee enhanced performance.

\begin{figure}[t]
    \centering
    \includegraphics[width=0.99\textwidth]{Styles/Figures/combined_heatmap_ver2.png}
    \caption{The heatmap visualization presents the \textbf{angular coverage} across all learned prompts through the 2nd, the 3rd, and the 4th round of training. The maximum angular coverage (MAC) monotonically increases from \textbf{67.7}\textdegree\:in the 2nd round to \textbf{75.95}\textdegree\:in the final round. The gradual reduction in rate of change in angular coverage towards the final round suggests that the model nearing convergence.}
    \label{fig:heatmaps}
    \vspace{-3ex}
\end{figure}
\begin{figure}[t]
    \centering
    \includegraphics[width=0.49\textwidth]{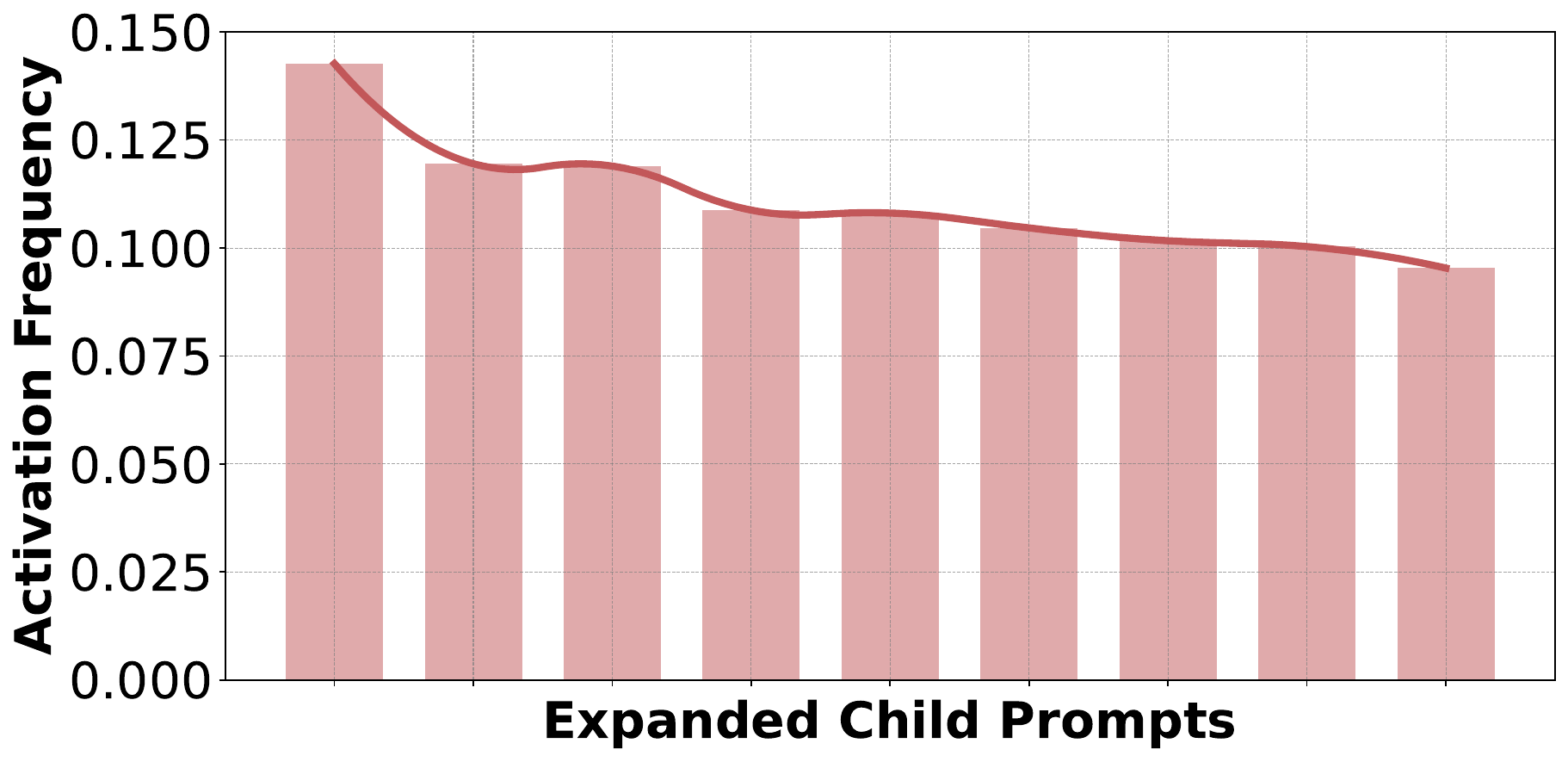}\hfill 
    \includegraphics[width=0.49\textwidth]{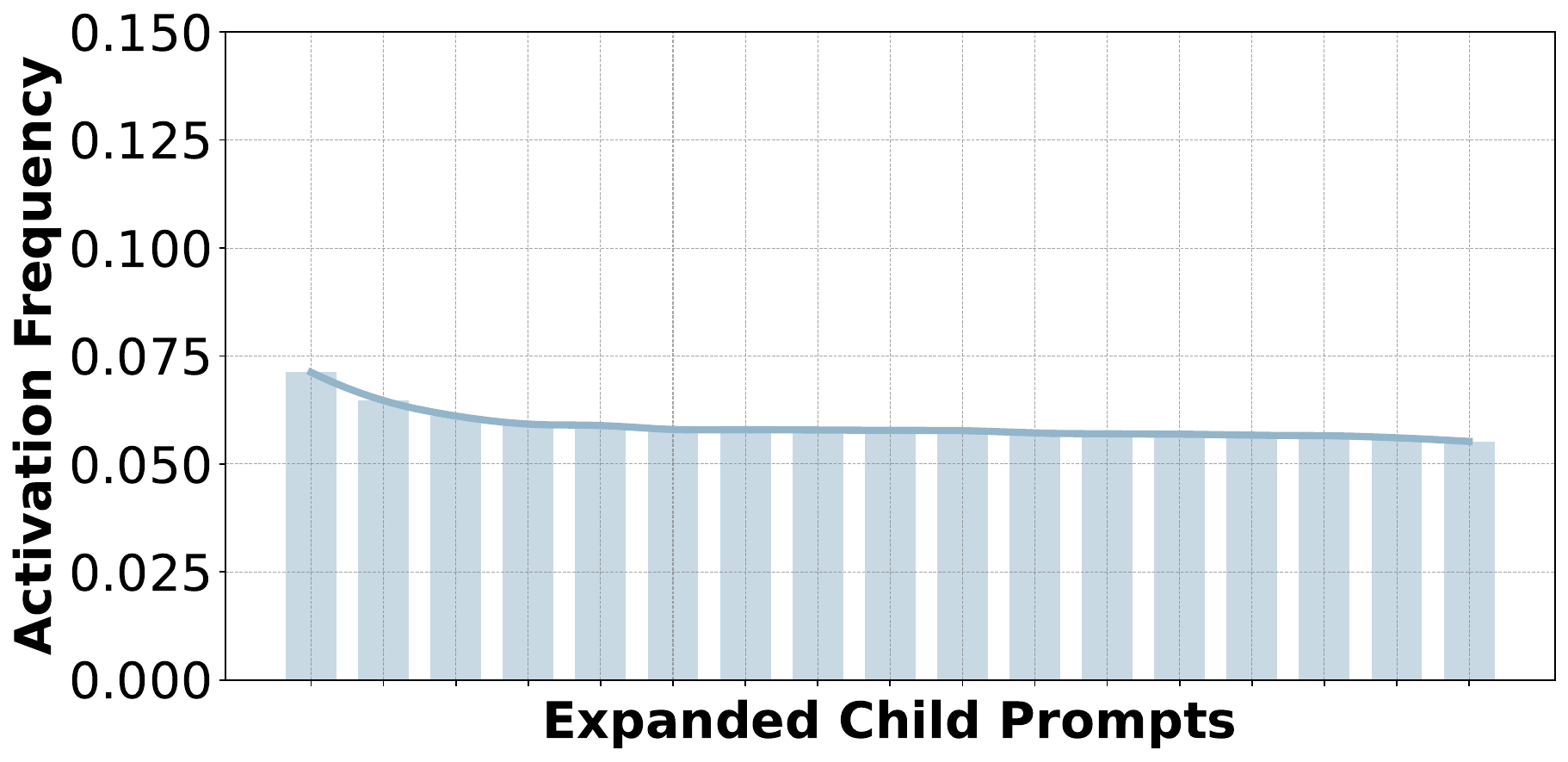}\vspace{-1ex}
    \caption{The distribution of \textbf{logit activation} of the learned prompts in the 2nd round (\textit{left}) and the 3rd round (\textit{right}). The prompt of the highest activation frequency is identified for further expansion.}
    \label{fig:activation}
\end{figure}

\textbf{Impact on Expansion Rounds and Angular Coverage.}
To substantiate our hypothesis that a higher maximum angular coverage (MAC) correlates with a broader spectrum of vocabularies, we computed the MAC using Equation \eqref{eq:mac}. The coverage results are visualized as heatmaps in Figure \ref{fig:heatmaps}. At the initial stage of expansion (leftmost heatmap), we observe that the prompts are quite uniformly distributed, with a mean coverage of 47.56\textdegree, This suggests that the prompts are actively exploring the embedding space to capture diverse semantics. As the expansion progresses to the third round (middle heatmap), the MAC increases from 67.78\textdegree~ to 75.70\textdegree. Specifically, row/col 7 (selected parent prompt) demonstrates the closest angular distances among the child prompts. This observation is crucial as it suggests that child prompts should not diverge excessively from the root semantics to maintain coherence. By the fourth round of expansion (rightmost heatmap), the pattern remains consistent with the third round. There is a reduced rate of change of MAC, achieving a maximum coverage of 75.95\textdegree and a mean coverage of 11.51\textdegree  among the child prompts. This plateau in MAC indicates that maximum semantic expansion has been reached, suggesting that the model is approaching convergence and further expansion may not be necessary. 

\textbf{The Distribution of Prompt Logit Activation.}
We previously established prompt logit activation frequency as an uncertainty measure to guide parent prompt selection for splitting. To investigate the dynamics of expanding highly uncertain parent prompts, we visualize the activation statistics (\textit{i.e.}, the frequency of logit activations) of tokens within the 2nd and 3rd expansion rounds. As illustrated in Figure \ref{fig:activation}, the distribution of these logits exhibits a long-tailed pattern, suggesting substantial uncertainty and numerous semantic overlaps among the mined semantics. The figure on the right demonstrates that, following the expansion of highly activated prompts, the distribution of child prompts becomes more uniform, suggesting the discovery of fine-grained semantics. These observations support our choice of uncertainty measure and verify the validity of DiPEx, indicating that expanding based on highly uncertain parent prompts effectively alleviates semantic ambiguity.

\begin{figure}[t]\vspace{-1ex}
    \includegraphics[width=\linewidth]{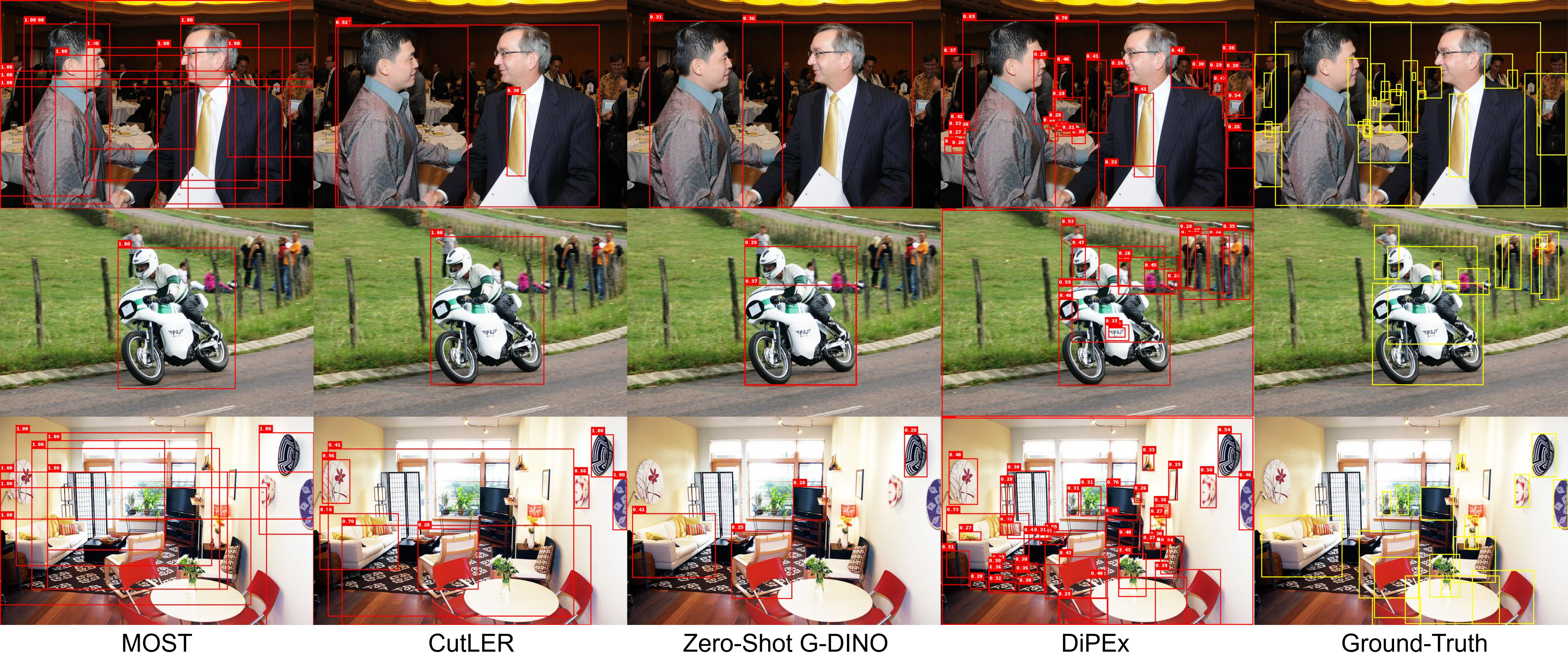}
    \vspace{-3ex}
    \caption{Visualization of the class-agnostic detection performance by baselines and the proposed DiPEx on MS-COCO \cite{DBLP:conf/eccv/LinMBHPRDZ14}. More visualizations are provided in Appendix (Figures \ref{fig:vis_box_more_1} and \ref{fig:vis_box_more_2}). \vspace{-3ex}}\label{fig:box_pred_vis}
\end{figure}

\noindent\textbf{Qualitative Study.}  In this section, we present visualized class-agnostic box predictions on images sampled from the MS-COCO dataset \cite{DBLP:conf/eccv/LinMBHPRDZ14}, as shown in Figure \ref{fig:box_pred_vis}. The proposed DiPEx method demonstrates a superior ability to detect more bounding boxes than all baseline methods, particularly for \textit{small objects}. For example, people in the distance (rows 1 and 3) and some bonsai (row 2) are missed by all baselines but successfully detected by DiPEx, showcasing its strong capability in localizing challenging small objects. For \textit{large objects}, such as a motorcycle (row 3) and two people shaking hands in the near distance (row 1), DiPEx localizes them with significantly higher confidence compared to the zero-shot predictions of Grounding DINO using the prompt “generic”. Additionally, DiPEx successfully identifies objects that are not annotated in the MS-COCO ground truth, such as plates (row 1), a pillowcase (row 2), and a frame on the wall (row 2). This highlights DiPEx's ability to identify a comprehensive set of class-agnostic objects, \textit{even} those missed in human annotations. 

\vspace{-2ex}
\section{Conclusion and Limitations}\vspace{-1ex}
This work introduces DiPEx, a novel self-supervised dispersing prompt expansion approach for class-agnostic object detection. We demonstrate through comprehensive experiments and analysis that DiPEx effectively detects a wide range of unseen objects of varying sizes and achieves broad vocabulary coverage. The progressively expanded prompt sets maintain good angular distances, promoting the formation of a semantic hierarchy and facilitating downstream detection tasks with a single inference pass. While the proposed DiPEx does not rely on box annotations, it requires self-training on the entire dataset for each round of prompt expansion, resulting in increased computational overhead. Additionally, some hyperparameters like temperature coefficients $\tau_p$, $\tau_c$ and learnable prompt length $K$, may require manual tuning for optimal performance. Future research directions include exploring methods to learn hierarchical prompts at once rather than through expansion. Extensive benchmarking on additional downstream tasks, such as open-vocabulary and open-world detection, is necessary to comprehensively validate the proposed approach.

\begin{ack}
This research is partially supported by the Australian Research Council (DE240100105, DP240101814, DP230101196)
\end{ack}
\bibliographystyle{plain}
\bibliography{main_arxiv}

%%%%%%%%%%%%%%%%%%%%%%%%%%%%%%%%%%%%%%%%%%%%%%%%%%%%%%%%%%%%
\newpage
\appendix

\section{Appendix / Supplemental Material}

This supplementary material includes a comprehensive overview of related work on class-agnostic object detection, vision-language models (VLMs), and prompt tuning. Additionally, we provide detailed descriptions of baselines, implementation details for both the baselines and the proposed method, and an extensive ablation study are provided. The ablation study analyzes the impact of pseudo-labeled supervision and the effect of the hyperparameter $\gamma$. Lastly, we present more comprehensive visualizations of class-agnostic box predictions.

\begin{figure}[t]\vspace{-1ex}
    {\includegraphics[width=0.99\textwidth]{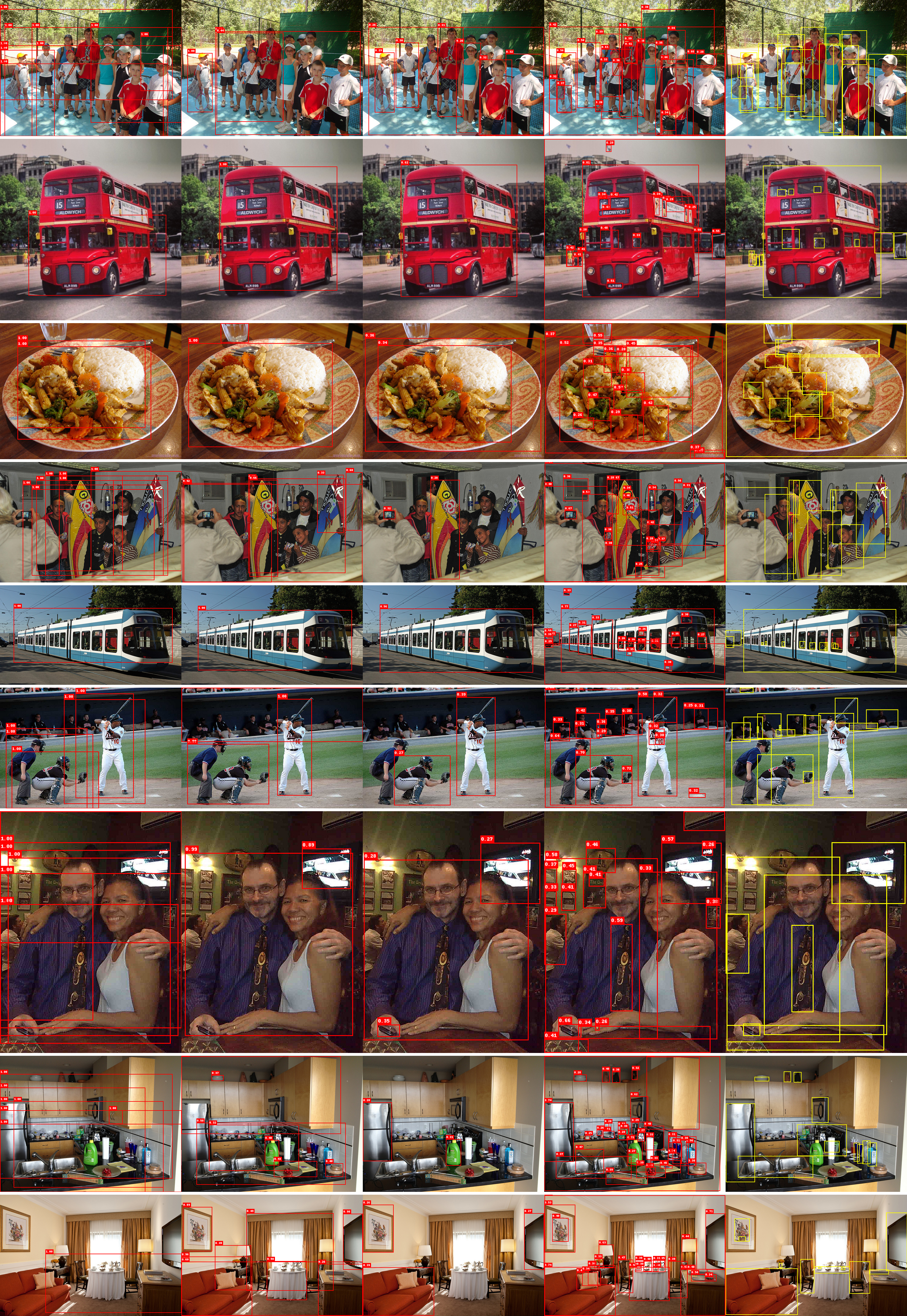}}    
    % {\includegraphics[width=0.99\textwidth]{Styles/Figures/Pred_vis/annotated_000000001000.jpg}}
    % {\includegraphics[width=0.99\textwidth]{Styles/Figures/Pred_vis/annotated_000000001584.jpg}}
    % {\includegraphics[width=0.99\textwidth]{Styles/Figures/Pred_vis/annotated_000000003845.jpg}}
    % {\includegraphics[width=0.99\textwidth]{Styles/Figures/Pred_vis/annotated_000000005193.jpg}}
    % {\includegraphics[width=0.99\textwidth]{Styles/Figures/Pred_vis/annotated_000000006040.jpg}}
    % {\includegraphics[width=0.99\textwidth]{Styles/Figures/Pred_vis/annotated_000000006471.jpg}}
    % {\includegraphics[width=0.99\textwidth]{Styles/Figures/Pred_vis/annotated_000000006763.jpg}}
    % {\includegraphics[width=0.99\textwidth]{Styles/Figures/Pred_vis/annotated_000000007574.jpg}}
    % {\includegraphics[width=0.99\textwidth]{Styles/Figures/Pred_vis/annotated_000000029596.jpg}}
    \vspace{-1ex}
    \caption{Additional visualizations of class-agnostic box predictions. Columns 1 -- 4 correspond to the following methods: MOST \cite{DBLP:conf/iccv/RambhatlaMCS23}, CutLER \cite{DBLP:conf/cvpr/0007GYM23}, zero-shot Grounding DINO [“generic”] \cite{DBLP:conf/naacl/DevlinCLT19}, and our proposed DiPEx, respectively. The final column presents human-annotated ground truth bounding boxes from the MS-COCO dataset \cite{DBLP:conf/eccv/LinMBHPRDZ14}. 
    \vspace{-2ex}} \label{fig:vis_box_more_1}
\end{figure}

\begin{figure}[t]\vspace{-1ex}
    {\includegraphics[width=0.99\textwidth]{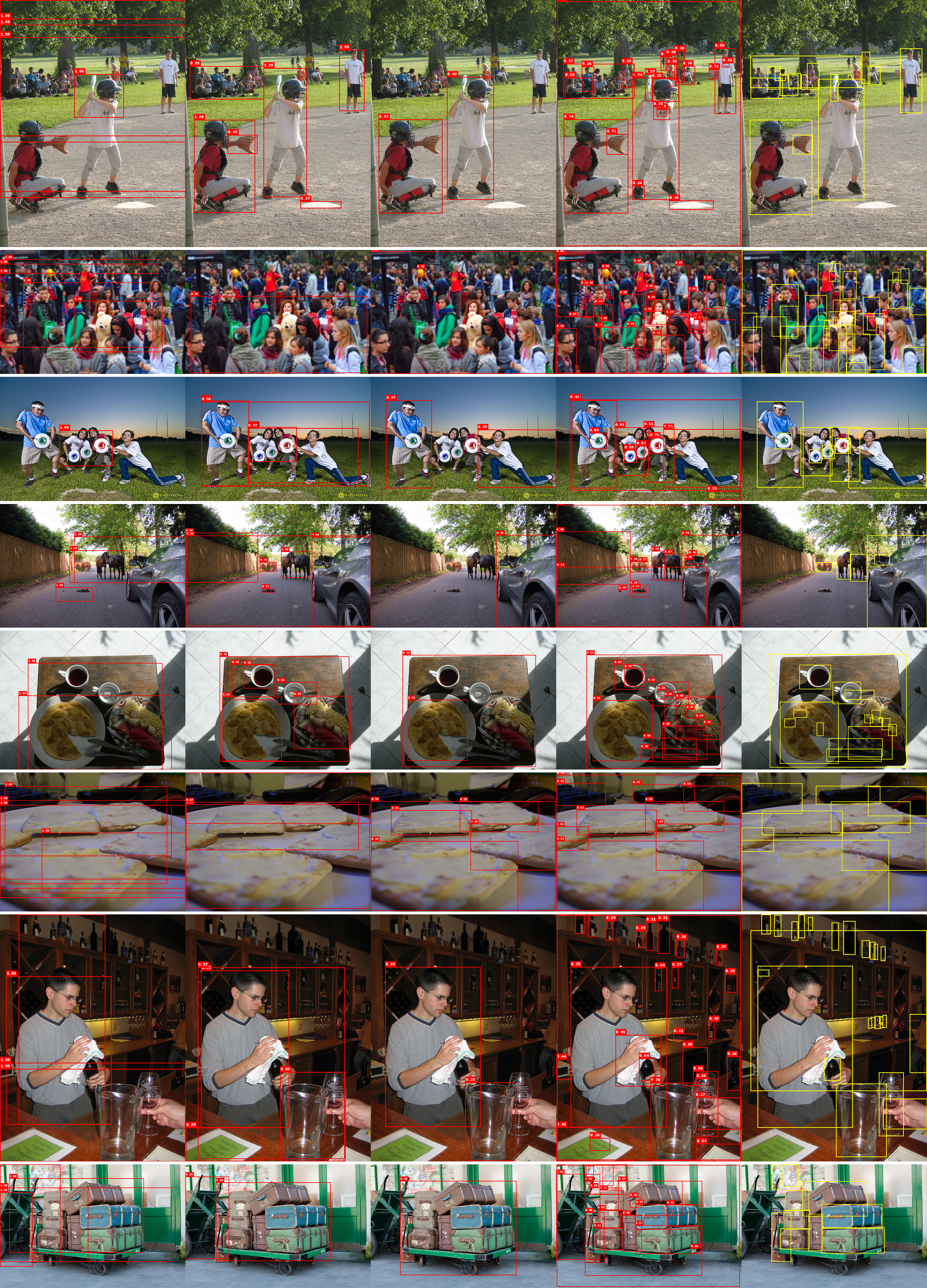}}
    % {\includegraphics[width=0.99\textwidth]{Styles/Figures/Pred_vis/annotated_000000012639.jpg}}
    % {\includegraphics[width=0.99\textwidth]{Styles/Figures/Pred_vis/annotated_000000012670.jpg}}
    % {\includegraphics[width=0.99\textwidth]{Styles/Figures/Pred_vis/annotated_000000013291.jpg}}
    % {\includegraphics[width=0.99\textwidth]{Styles/Figures/Pred_vis/annotated_000000017178.jpg}}
    % {\includegraphics[width=0.99\textwidth]{Styles/Figures/Pred_vis/annotated_000000017714.jpg}}
    % {\includegraphics[width=0.99\textwidth]{Styles/Figures/Pred_vis/annotated_000000021503.jpg}}
    % {\includegraphics[width=0.99\textwidth]{Styles/Figures/Pred_vis/annotated_000000025394.jpg}}
    % {\includegraphics[width=0.99\textwidth]{Styles/Figures/Pred_vis/annotated_000000026941.jpg}}
    \vspace{-1ex}
    \caption{Additional visualizations of class-agnostic box predictions. Columns 1 -- 4 correspond to the following methods: MOST \cite{DBLP:conf/iccv/RambhatlaMCS23}, CutLER \cite{DBLP:conf/cvpr/0007GYM23}, zero-shot Grounding DINO [“generic”] \cite{DBLP:conf/naacl/DevlinCLT19}, and our proposed DiPEx, respectively. The final column presents human-annotated ground truth bounding boxes from the MS-COCO dataset \cite{DBLP:conf/eccv/LinMBHPRDZ14}. 
    \vspace{-2ex}} \label{fig:vis_box_more_2}
\end{figure}

\begin{itemize}
    \item \textbf{Section \ref{sec:related_work}:} Related Work
    % \item \textbf{Sec. \ref{sec:case_study}:} Case Study - Confidence Diminishing
    \item \textbf{Section \ref{sec:impl_details}:} Baselines and Implementation Details
    \item \textbf{Section \ref{sec:more_ablation}:} More Ablation Studies
    % \item \textbf{Algorithm \ref{alg:dipex}:} Detailed Algorithm of DiPEx
    \item \textbf{Figures \ref{fig:vis_box_more_1} and \ref{fig:vis_box_more_2}:} Additional Visualizations of Class-Agnostic Box Predictions

\end{itemize}

\subsection{Related Work}\label{sec:related_work}
\textbf{Class-Agnostic Object Detection.} Traditional bottom-up approaches \citep{DBLP:journals/ijcv/UijlingsSGS13, DBLP:conf/eccv/KrahenbuhlK14, DBLP:conf/eccv/ZitnickD14, DBLP:conf/nips/PinheiroCD15, DBLP:journals/pami/Pont-TusetABMM17, DBLP:conf/cvpr/ChengZLT14} for region proposal generation, often grapple with the constraints precision, despite high recall rates, reducing their scalability for general use in diverse environments. Recent breakthroughs in ViTs \citep{DBLP:conf/iccv/CaronTMJMBJ21, DBLP:conf/iclr/DosovitskiyB0WZ21, DBLP:journals/corr/abs-2304-07193} have enabled scaling up to massive datasets for self-supervised learning, extracting both local and global semantically meaningful features. This has led to numerous methods in unsupervised object discovery and localization. LOST \cite{DBLP:conf/bmvc/SimeoniPVRGBPMP21} is an early application, using a patch similarity graph and an inverse degree map to identify seed patches and extract bounding boxes. TokenCut \cite{DBLP:conf/cvpr/Wang0H0CV22} constructs an undirected graph with image tokens as nodes, applying the normalized cut algorithm \cite{DBLP:journals/pami/ShiM00} for foreground-background segregation. MOVE \cite{DBLP:conf/cvpr/Melas-Kyriazi0L22} builds on LOST by employing deep spectral bipartitioning, offering a more principled and effective approach. However, both LOST \cite{DBLP:conf/bmvc/SimeoniPVRGBPMP21} and TokenCut \cite{DBLP:conf/cvpr/Wang0H0CV22} are limited to detecting a single object per image. MOST \cite{DBLP:conf/iccv/RambhatlaMCS23} addresses this limitation by using entropy-based box analysis (EBA) to segregate foreground tokens. Nevertheless, their performance remains \textit{sub-optimal}, constrained by their limited capacity to generalize across diverse object categories. Closest to our work is MAVL \cite{DBLP:conf/eccv/MaazR0KA022}, where they develop an MViT with late fusion strategy and use generic text queries like “\texttt{all objects}” to locate objects. However, their framework requires full supervision and multiple inferences with different textual prompts, yet still falls short of achieving optimal performance. In contrast, our approach eliminates the need for labels and achieves SOTA performance with one-pass inference with the non-overlapping prompts.

\textbf{VLMs and Prompt Tuning.} Recent advances in VLMs \cite{DBLP:conf/icml/RadfordKHRGASAM21, DBLP:conf/icml/JiaYXCPPLSLD21,DBLP:conf/iclr/YaoHHLNXLLJX22} which are pretrained on expansive image-text pairs have demonstrated significant potential in learning generic concepts. HierKD \cite{DBLP:conf/cvpr/MaLGLCWZH22} introduces global language-to-visual knowledge distillation modules, which align global-level image representations with caption embeddings through contrastive loss. OV-DETR \cite{DBLP:conf/cvpr/ZareianRHC21} pioneered the extension of the DETR framework to an open-vocabulary setting by integrating a conditional binary matching mechanism. GLIP \cite{DBLP:conf/cvpr/LiZZYLZWYZHCG22} converted object detection into a grounding task, utilizing additional data to align phrase and region semantics. Recently, Grounding DINO \cite{liu2023grounding} introduced a dual-encoder-single-encoder framework to integrate object detection and visual grounding within a unified architecture. Similarly, T-Rex2 \cite{DBLP:journals/corr/abs-2403-14610} synergizes text and visual prompts through contrastive learning,leading to state-of-the-art performance in out-of-distribution object detection. Nonetheless, the effectiveness of VLMs is heavily influenced by the textual cues they are conditioned on, and efficiently adapting them to specific downstream applications remains a substantial challenge as manually engineering optimal prompts can often entail considerable effort and resources \cite{DBLP:journals/corr/abs-2304-00685}. Prompt tuning is a simple yet effective solution to adapt models to specific tasks by optimizing a small number of soft prompts in an end-to-end manner while keeping the original model's parameters frozen. The pioneering work of CoOp \cite{zhou2022coop} introduced context optimization by fine-tuning CLIP using learnable tokens. However, CoOp's generalizability was constrained, a limitation later addressed by CoCoOp \cite{zhou2022cocoop}, which conditioning input tokens on image embeddings. MaPLe \cite{DBLP:conf/cvpr/KhattakR0KK23} advanced this by introducing a multi-modal prompting technique to overcome the limitations of uni-modal prompting methods. ProDA \cite{DBLP:conf/cvpr/LuLZL022} further innovated by learning a distribution of diverse prompts and employing a Gaussian model to capture visual variations. Despite these advancements, an inherent limitation persists across these methods: they all require full supervision. UPL \cite{DBLP:journals/corr/abs-2204-03649} first proposed unsupervised prompt learning for image recognition task, POUF \cite{DBLP:conf/icml/TanwisuthZZHZ23} later introduced a similar self-prompting mechanism to minimize entropy using optimal transport. However, these methods have yet to be adapted for the object detection domain. To our knowledge, UPT \cite{DBLP:conf/iccv/HeCCYXLQZ23} is the only existing work that optimizes prompts using dual complementary teaching specifically for object detection tasks. Our work, DiPEx, represents the first endeavor to apply prompt learning to class-agnostic object detection through a self-training approach.

\subsection{Baselines and Implementation Details}\label{sec:impl_details}
\noindent \textbf{Baselines.} We compare the proposed approach with fourteen baselines: 1) bottom-up selective search \cite{DBLP:journals/ijcv/UijlingsSGS13} that slides windows of different sizes to locate objects, 2) UP-DETR \cite{DBLP:conf/cvpr/DaiCLC21}, an unsupervised pre-training method for OD that can be fine-tuned to detect class-agnostic objects. 3) DETReg \cite{DBLP:conf/cvpr/BarWKRHCRDG22}, which learns to localize objects and encode an object’s properties during unsupervised pre-training, 4) MOST \cite{DBLP:conf/iccv/RambhatlaMCS23}, a multiple objects localizer based on patch correlations without any training, 5) FreeSOLO \cite{DBLP:conf/cvpr/WangYMKASA22}, which unifies pixel grouping, object localization and feature pre-training in a fully self-supervised manner, 6) Exemplar-FreeSOLO \cite{DBLP:conf/cvpr/IshtiakEG23}, an improved approach based on FreeSOLO through exemplar knowledge extraction, 7) CutLER \cite{DBLP:conf/cvpr/0007GYM23}, an unsupervised object detection method by encouraging the detector to explore objects missed in extracted coarse masks, 8) HASSOD \cite{DBLP:conf/nips/CaoJGW23}, a clustering strategy that groups regions into object masks based on self-supervised features, 9) CoOp \cite{zhou2022coop} and 10) CoCoOp \cite{zhou2022cocoop}, prompting techniques that utilize learnable vectors to model a prompt's context words, enabling zero-shot transfer to class-agnostic detection, 11) segment anything model (SAM) \cite{DBLP:conf/iccv/KirillovMRMRGXW23}, a foundational model trained on 1 billion masks and 11 million images such that can perform zero-shot transfer to the class-agnostic OD task. For OOD-OD task, we further compare three baseline methods: 12) VOS \cite{DBLP:conf/iclr/DuWCL22} that regularizes the model's decision boundary between known and unknown classes by training with generated virtual outliers. 13) PROB \cite{DBLP:conf/cvpr/ZoharWY23} which utilizes a multivariate Gaussian distribution to learn objectness probability to separate known and unknown objects, 14) UnSniffer \cite{DBLP:conf/cvpr/LiangXLZM23}, which similarly introduces an object confidence, derived from learning known objects with varying degrees of overlap. 

\noindent \textbf{Implementation Details.} Our code is developed on the Open Grounding-DINO framework \cite{Open_Grounding_Dino}, and operates on a single NVIDIA RTX A6000 GPU with 48 GB of memory. For our experiments, we choose a batch size of 8 for training, and set hyperparameter $\gamma=0.1$, $\tau_{p}=0.1$, $\tau_{c}=0.1$, $\theta=\pm15^{\circ}$, $K=9$, $L=3$, and  while adopting all remaining hyperparameters from the Open Grounding-DINO codebase. We empirically set the $\Tau_{\alpha}=75^{\circ}$ as our threshold for expansion termination. The original implementation of CoOP was developed for image classification tasks based on CLIP and supervised contrastive learning. We extend CoOP to class-agnostic object detection using pseudo labeling-based self-training, which remains consistent with our approach. All the implementation code and configurations files are provided in supplementary materials and will be publicly released upon acceptance of this work.

\subsection{More Ablation Studies} \label{sec:more_ablation}
\noindent \textbf{Pseudo-labels Construction}
For pseudo-labeling, we utilize off-the-shelf Grounding DINO with a “generic” text prompt, which demonstrates considerable zero-shot performance, as illustrated in our pilot study. Additionally, we generate pseudo-boxes by concatenating all 25 beginner nouns from WordNet \cite{fellbaum1998wordnet}. We then merge the predictions from these two queries and apply Soft-NMS to eliminate overlaps. In the following Section \ref{sec:pl_impact}, we also investigate the performance of DiPEx alongside other prompt-tuning methods on the quality of pseudo-labels.

\begin{table*}[t]
\centering 
\caption{The impact on pseudo-labeled supervision on MS-COCO \cite{DBLP:conf/eccv/LinMBHPRDZ14} dataset, when applying different pseudo-labels queried on Grounding DINO using different textual cues. In the main paper, we report the performance of DiPEx using merged pseudo labels for the first round of training.}
\resizebox{0.99\linewidth}{!}{%
\begin{tabular}{l c c c c c c c c c}
\toprule
\multirow{1}{*}{Method} & AR$_{100}$ & AR@S & AR@M & AR@L &AP &AP@S &AP@M &AP@L\\
\midrule 
Grounding DINO @ [“generic”] & 44.1 & 17.7 & 51.6 & 80.0 & 28.3 & 11.4 & 33.0 & 56.5 \\
Grounding DINO @ [25 nouns] & 40.5 & 16.0 & 46.6 & 75.0 & 12.1 & 4.4 & 12.0 & 26.7 \\
Grounding DINO @ merged & 51.9 & 24.8 & 61.6 & 85.8 & 19.1 & 7.6 & 19.5 & 42.0 \\
\midrule
\rowcolor{LightCyan!40}\textbf{DiPEx} @ [“generic”] & \textbf{65.5} & \textbf{42.7} & \textbf{76.2} & \textbf{90.3} & \textbf{37.0} & \textbf{20.5} & \textbf{43.3} & 62.4 \\
\rowcolor{LightCyan!40}\textbf{DiPEx} @ [25 nouns] & 46.6 & 18.5 & 55.7 & 83.3 & 13.2 & 4.0 & 13.3 & 30.7 \\
\rowcolor{LightCyan!40}\textbf{DiPEx} @ merged & 63.2 & 39.2 & 74.3 & 89.8 & 35.9 & 16.4 & 39.7 & \textbf{63.8}    \\
\bottomrule
\end{tabular}}
\label{table:pl_impact}
\end{table*}

\noindent \textbf{Impact on Pseudo-labeled Supervision.} \label{sec:pl_impact} In this section, we investigate how the quality of pseudo-labels used for self-training impacts DiPEx's performance, given our reliance on these training samples. Specifically, we generate pseudo-labels by querying the off-the-shelf Grounding Dino model with three different approaches: 1) using a “generic” text prompt, 2). the 25 beginner nouns from WordNet \cite{fellbaum1998wordnet}, and a combination of both. As shown in Table \ref{table:pl_impact}, the “generic” text prompt alone demonstrated considerable performance. However, we observed an improvement in Average Recall (AR) when merging the predictions generated by “generic” with the 25 beginner nouns, leading us to this study. Consequently, we use these pseudo-labels to self-train our model.

\begin{wrapfigure}{r}{0.5\linewidth}\vspace{-2ex}
  \begin{center}
\vspace{-2ex}
    \includegraphics[width=0.85\linewidth]{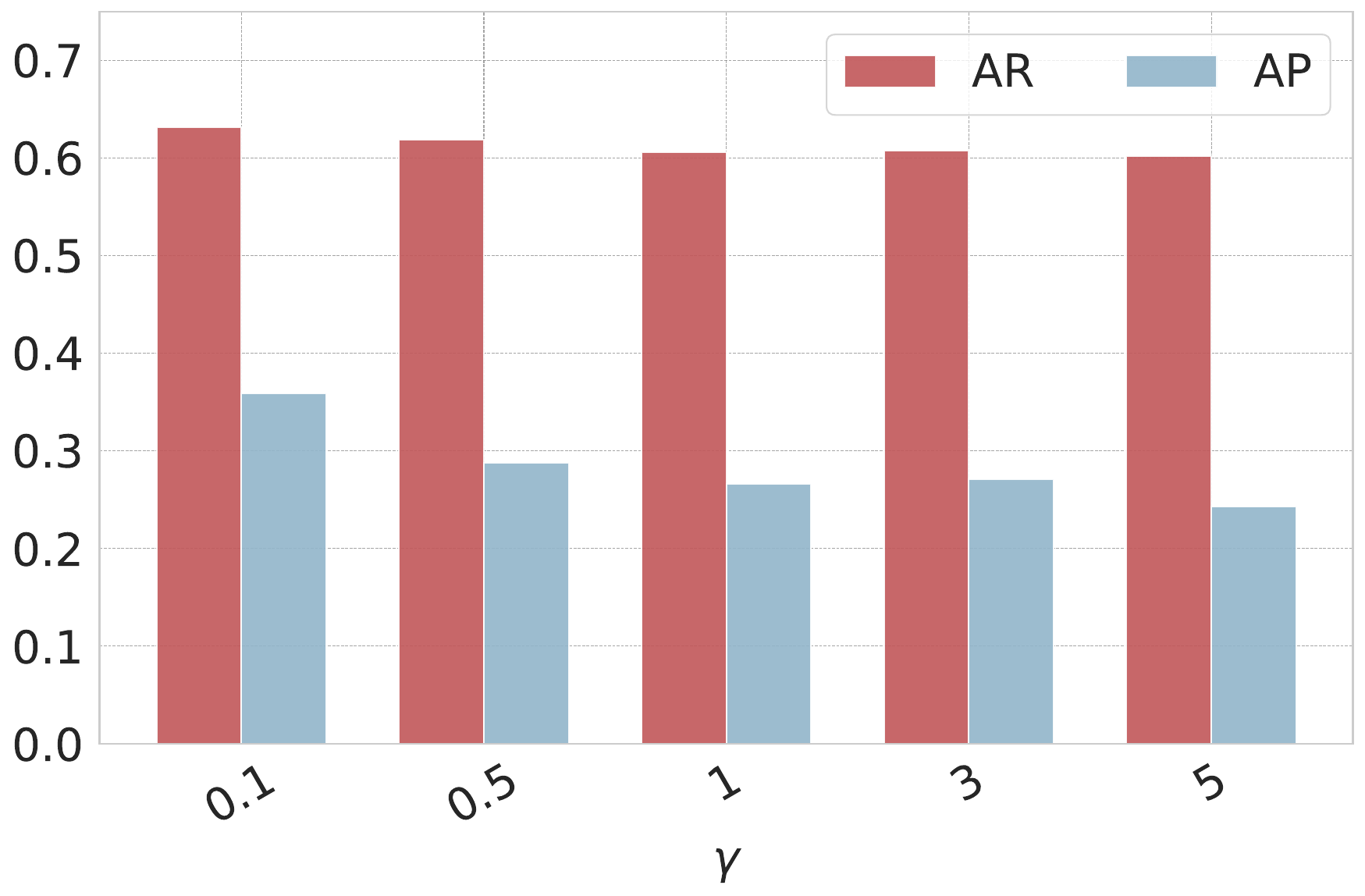}
  \end{center}
\vspace{-2ex}
  \caption{Study of Loss Coefficient $\gamma$ }\label{fig:method_workflow} \vspace{-2ex}
\end{wrapfigure}

\noindent \textbf{Effect of Loss Coefficient $\gamma$.}
To effectively separate child prompts while maintaining semantic coherence between parent and child prompts, selecting an appropriate $\gamma$ is essential. As illustrated in the bar plot below, a moderate $\gamma$ value typically yields optimal results. In contrast, a larger value (e.g., $\gamma = 5$) causes child prompts to diverge more significantly, distorting semantic integrity and potentially leading to over-regularization of the model.

\clearpage
\newpage
\end{document}

%% file: math_extension.tex
\newcommand{\eat}[1]{}

\usepackage{booktabs}
\usepackage{multirow}
\usepackage{siunitx}
\usepackage{bm}

\usepackage{mathtools}
\usepackage{bbm}
\DeclarePairedDelimiterX{\infdivx}[2]{(}{)}{%
  #1\;\delimsize\|\;#2%
}

\newcommand{\Tau}{\mathcal{T}}

\usepackage{stmaryrd}

\makeatletter
\newcommand{\xMapsto}[2][]{\ext@arrow 0599{\Mapstofill@}{#1}{#2}}
\def\Mapstofill@{\arrowfill@{\Mapstochar\Relbar}\Relbar\Rightarrow}
\makeatother

\usepackage{enumitem,kantlipsum}
\usepackage{algorithm}

\usepackage{algpseudocode}

\usepackage{mathrsfs}

\algnewcommand{\Initialize}[1]{%
  \State \textbf{Initialize:}
  \Statex \hspace*{\algorithmicindent}\parbox[t]{.8\linewidth}{\raggedright #1}
}
\algnewcommand{\Inputs}[1]{%
  \State \textbf{Inputs:}
  \Statex \hspace*{\algorithmicindent}\parbox[t]{.8\linewidth}{\raggedright #1}
}
\algnewcommand{\Outputs}[1]{%
  \State \textbf{Outputs:}
  \Statex \hspace*{\algorithmicindent}\parbox[t]{.8\linewidth}{\raggedright #1}
}

\definecolor{seal}{HTML}{B711AC}
\definecolor{maroon}{cmyk}{0,0.87,0.68,0.32}

\usepackage{tikz}
\usepackage{balance}